\documentclass[10pt,journal,compsoc]{IEEEtran}
\ifCLASSOPTIONcompsoc
  \usepackage[nocompress]{cite}
\else
  \usepackage{cite}
\fi
\ifCLASSINFOpdf
   \usepackage[pdftex]{graphicx}
\else
\fi
\usepackage{nicefrac}
\usepackage{subfigure}
\usepackage[table]{xcolor}
\usepackage{multirow}
\usepackage{amssymb}
\usepackage{amsfonts}
\usepackage{booktabs}
\usepackage{hyperref}
\usepackage{adjustbox,lipsum}
\usepackage{footnote}
\usepackage{tablefootnote}

\newcommand{\hRevise}{black}
\newcommand{\hReviseMinor}{black}
\usepackage{amsmath}
\usepackage{algorithmic}

\usepackage{array}

\usepackage{makecell}

\usepackage{url}

\begin{document}
%

\title{Spherical Kernel for Efficient \\Graph Convolution on 3D Point Clouds}

%
%
%
%

\author{Huan~Lei,
        Naveed~Akhtar,
        and~Ajmal~Mian
\IEEEcompsocitemizethanks{\IEEEcompsocthanksitem The authors are with the Department of Computer Science and
Software Engineering, The University of Western Australia, 35 Stirling
Highway, Crawley, Western Australia, 6009.
E-mail: huan.lei@research.uwa.edu.au, naveed.akhtar@uwa.edu.au, 
ajmal.mian@uwa.edu.au.
}
}

%
%

\markboth{Journal of \LaTeX\ Class Files,~Vol.~x, No.~x, September~2019}%
{Shell \MakeLowercase{\textit{et al.}}: Bare Demo of IEEEtran.cls for Computer Society Journals}
\IEEEtitleabstractindextext{%
\begin{abstract}
We propose a spherical kernel for efficient graph convolution of 3D point clouds. 
Our metric-based kernels systematically quantize the local 3D space 
to identify distinctive geometric relationships in the data. Similar to the regular grid CNN kernels, the spherical kernel maintains translation-invariance and asymmetry properties, where the former guarantees weight sharing among similar local structures in the  data and the latter facilitates fine geometric learning. 
The proposed kernel is applied to graph neural networks without edge-dependent filter generation, making it computationally attractive for large point clouds. 
In our graph networks, each vertex is associated with a single point location and edges connect the neighborhood points within a defined range. The graph gets coarsened in the network with farthest point sampling. 
Analogous to the standard CNNs, we define pooling and unpooling operations for our network. 
We demonstrate the effectiveness of the proposed spherical kernel with graph neural networks for point cloud classification and semantic segmentation  using ModelNet, ShapeNet, RueMonge2014, ScanNet and S3DIS datasets.
The source code and the trained models can be downloaded from  \href{https://github.com/hlei-ziyan/SPH3D-GCN}{https://github.com/hlei-ziyan/SPH3D-GCN}.
\end{abstract}

}

\maketitle

\IEEEdisplaynontitleabstractindextext

%
\IEEEpeerreviewmaketitle

\IEEEraisesectionheading{\section{Introduction}\label{sec:introduction}}

%
%
%
%
\IEEEPARstart{C}{onvolutional} 
neural networks (CNNs) \cite{lecun1998gradient} are known for accurately solving a wide range of Computer Vision problems. 
Classification \cite{krizhevsky2012imagenet,simonyan2014very,Szegedy2015googLeNet,he2016deep}, image segmentation \cite{long2015fully,ronneberger2015u,badrinarayanan2017segnet}, object detection \cite{redmon2016you,liu2016ssd,ren2015faster}, and  face recognition \cite{schroff2015facenet,gilani2018dense} are just a few examples of the tasks for which CNNs have recently become the default modelling technique. 
The success of CNNs is mainly attributed to their impressive representational prowess. 
However, their representation is only amenable to the data defined over regular grids, e.g.~pixel arrays of images and videos. This is problematic for applications where the data is inherently irregular \cite{bronstein2017geometric},  e.g.~3D Vision, Computer Graphics and Social Networks.

In particular, point clouds produced by 3D vision scanners (e.g. LiDAR, Matterport) are highly irregular. 
Recent years have seen a surge of interest in deep learning for 3D vision due to self-driving vehicles. This has also resulted in    
multiple public repositories of 3D point clouds \cite{chang2015shapenet,yi2016scalable}, \cite{armeni20163d}, \cite{dai2017scannet}, \cite{hackel2017semantic3d}.  
Early attempts of exploiting CNNs for point clouds applied regular grid transformation (e.g. voxel grids \cite{wu20153d,maturana2015voxnet}, multi-view images \cite{su2015multi}) to point clouds for processing them with 3D-CNNs or enhanced 2D-CNNs \cite{simonyan2014very,he2016deep}.   
However, this line of action does not fully exploit the sparse nature of point clouds, leading to unnecessarily large memory footprint and computational overhead of the methods.
Riegler \emph{et al.}~\cite{riegler2017octnet} addressed the memory issue in dense 3D-CNNs with an octree-based network, termed OctNet. However, the redundant computations over empty spaces still remains a discrepancy of OctNet. 

Computational graphs are able to capitalize on the sparse nature of point clouds much better than volumetric or multi-view representations.
However, designing effective modules such as convolution, pooling and unpooling layers,  becomes a major challenge for the graph based convolutional networks.
These modules are expected to perform point  operations analogous to the pixel operations of  CNNs, albeit for 
irregular data.
Earlier instances of such modules exist in theoretical works~\cite{bruna2013spectral,defferrard2016convolutional,kipf2017semi}, which can be exploited to form Graph Convolutional Networks (GCNs)~\cite{kipf2017semi}.  
Nevertheless, these primitive GCNs are yet to be seen as a viable solution for point cloud processing due to their inability to effectively handle real-world point clouds. 

\begin{figure*}[t]
  \centering
  \includegraphics[width=0.96\textwidth]{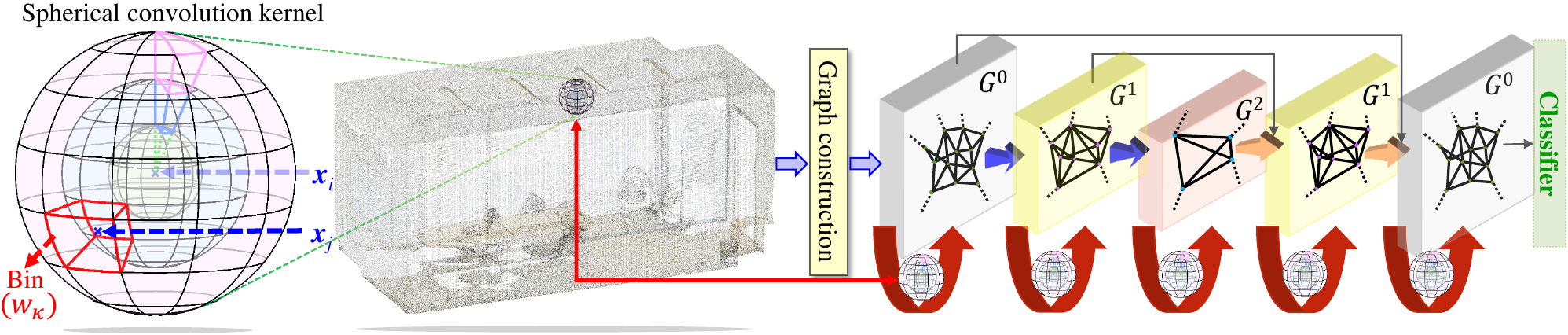}
  \caption{
  The proposed spherical convolutional kernel systematically splits the space around a point ${\bf x}_i$ into multiple volumetric bins. For the $j^{\text{th}}$ neighboring point ${\bf x}_j$, it determines its relevant bin and uses the weight $w_{\kappa}$ for that bin to compute the activation. 
  The kernel is employed with graph based networks that directly process raw point clouds using a pyramid of graph representations. This is a simplified U-Net-like\cite{badrinarayanan2017segnet} architecture for semantic segmentation that coarsens the input graph $G^0$ into $G^1$ with pooling, and latter uses unpooling for expansion. In the network, the location of a point identifies a graph vertex and point neighbourhood decides the graph edges. Our network allows convolutional blocks with consecutive applications of the proposed kernels for more effective representation learning.}
  \label{fig:intro_arch}
  \vspace{-2mm}
\end{figure*}

Based on the convolution operation, GCNs can be divided into two groups, namely; the \textit{spectral} networks \cite{bruna2013spectral,defferrard2016convolutional,kipf2017semi,yi2017syncspeccnn} and the \textit{spatial} networks~\cite{simonovsky2017dynamic,wang2018dynamic,wang2019graph,wu2019pointconv,lei2019octree}. The former perform convolutions using the graph Laplacian and adjacency matrices, whereas the latter perform convolutions directly in the spatial domain. 
For the spectral networks, careful alignment of the graph Laplacians of different samples is necessary~\cite{yi2017syncspeccnn}. This is not easily achieved for the real-world point clouds. 
Consequently, the spatial networks are generally considered more attractive than the spectral networks in practical applications.


The spatial GCNs are challenged by the unavailability of discrete convolutional kernels in the 3D metric space. 
To circumvent the problem,
mini-networks~\cite{simonovsky2017dynamic,wang2018dynamic,wu2019pointconv} are often applied to dynamically generate edge-wise filters. This incurs significant computational overhead, which can be avoided in the case of discrete kernels. 
However,  design and application of discrete kernels in this context is not straight forward.
Beside effective discretization of the metric space, the kernel application must exhibit the properties of (a) translation-invariance that allows identification of similar local structures in the data, and (b) asymmetry for vertex pair processing to ensure that the overall representation remains compact.   


Owing to the intricate requirements of discrete kernels for irregular data, many existing networks  altogether avoid the convolution operation for point cloud processing~\cite{qi2017pointnet},\cite{klokov2017escape, qi2017pointnetplusplus, li2018so}. 
Although these techniques report decent performance on benchmark datasets, they do not contribute towards harnessing the power of convolutional networks for point clouds.
PointCNN~\cite{li2018pointcnn} is a notable exception that uses a convolutional kernel for point cloud 
processing. However, its kernel is again defined  using mini-networks, incurring high computational cost. Moreover, it is sensitive to the order of the neighborhood points, implicating that the underlying operation is not permutation-invariant, which is not a desired kernel property for point clouds.

In this work, we introduce a discrete metric-based \textit{spherical} convolutional kernel that  systematically partitions a 3D region into multiple volumetric bins as shown in Fig.~\ref{fig:intro_arch}.
The kernel is directly applied to point clouds for convolution. 
Each bin of the kernel specifies learnable parameters to convolve the points falling in it.
The convolution defined by our kernel   preserves the properties of translation-invariance, asymmetry, as well as permutation-invariance. 
The proposed kernel is applied to point clouds using Graph Networks.
To that end, we construct the networks with the help of range search~\cite{preparata2012computational} and farthest point sampling~\cite{qi2017pointnetplusplus}. The former  defines edges of the underlying graph, whereas the latter coarsens the graph as we go deeper into the network layers.  
We also define pooling and unpooling modules for our graph networks to downsample and upsample the vertex features. 
The novel convolutional kernel and its application to graph networks are   thoroughly evaluated for the tasks of 3D point cloud classification and semantic segmentation. We achieve highly competitive performance on a wide range of benchmark datasets, including  ModelNet~\cite{wu20153d}, ShapeNet \cite{yi2016scalable}, RueMonge2014 \cite{riemenschneider2014learning}, ScanNet \cite{dai2017scannet} and S3DIS \cite{armeni20163d}.
Owing to the proposed kernel, the resulting graph networks are found to be efficient in both memory and computation.
This leads to fast training and inference on high resolution point clouds. 

This work is a significant extension of our preliminary findings presented in IEEE CVPR 2019~\cite{lei2019octree}. 
Below, we summarize the major directions along which the technique is extended beyond the preliminary work. 
\begin{itemize}
\item \textbf{Separable convolution.} 
We perform the depth-wise and point-wise convolution operation separately in this work rather than simultaneously 
as in \cite{lei2019octree}. 
The separable convolution strategy is inspired by Xception~\cite{chollet2017xception}, and significantly reduces the number of network parameters and computational cost. 
\item \textbf{Graph architecture.}
Instead of the octree-guided network of \cite{lei2019octree}, we use a more flexible graph-based technique to design our network architectures.
This allows us to exploit convolution blocks and  define pooling/unpooling operations independent of convolution. 
In contrast to the convolution-based down/upsampling, specialized modules for these operations are highly desirable for processing large point clouds. 
Moreover, this strategy also brings our network architectures closer to the standard CNNs.   
\item \textbf{Comprehensive evaluation on real-world  data}.
Compared to the preliminary work~\cite{lei2019octree}, we present a more thorough evaluation on real-world data. Highlights include  4.2\% performance gain over \cite{lei2019octree} for the RueMonge2014 dataset, and comprehensive evaluation on two additional datasets, ScanNet and S3DIS.
The presented results ascertain the computational efficiency of our technique with highly competitive performance on the popular benchmarks.
\item \textbf{Tensorflow implementation.} 
While~\cite{lei2019octree} was implemented in Matconvnet, with this article, we  release cuda implementations of the spherical convolution and the pooling/unpooling operations for Tensorflow. The source code is available on Github (\href{https://github.com/hlei-ziyan/SPH3D-GCN}{https://github.com/hlei-ziyan/SPH3D-GCN}) 
for the broader research community.
\end{itemize}

\section{Related Work}
\label{sec:RW}
PointNet~\cite{qi2017pointnet} is one of the first techniques to directly process point clouds with deep networks. It uses the $xyz$ coordinates of points as input features. The network learns point-wise features with shared MLPs, and extracts a global feature with max pooling. One limitation of this technique is that it does not explore the geometric context of points in representation learning. PointNet++~\cite{qi2017pointnetplusplus} addresses that by applying max-pooling to the local regions hierarchically.  
However, both networks must rely on max-pooling to aggregate any context information without convolution.

SO-Net \cite{li2018so} builds an $m\times m$ rectangular map from the point cloud, and hierarchically learns node-wise features within the map using mini-PointNet. However, similar to the original PointNet, it also fails to exploit any convolution modules.
KCNet \cite{shen2018mining} learns features with kernel correlation between the local neighboring points and a template of learnable points. This can be optimized in a  training session similar to convolutional kernels. In contrast to the image-like map used by the SO-Net, KCNet is based on graph representation.
Kd-network~\cite{klokov2017escape} is a  prominent contribution that processes  point clouds with tree structure based networks. This technique also uses point coordinates as the input and computes the feature of a parent node by concatenating the features of its children in a balanced tree. 
{\color{\hRevise}Instead of using $xyz$ coordinates, ShapeContextNet \cite{xie2018attentional} computes hand-crafted shape context descriptors  \cite{belongie2002shape} for each point in the point cloud, and explores them as input features for PoinetNet-like architecture.}
Despite their varied network architecture construction, none of the above methods contribute towards developing convolutional networks for point clouds. 
Approaches that advance research in that direction can be divided into two broad categories, discussed below. 

\vspace{-2mm}
\subsection{3D Convolutional Neural Networks}
At the advent of 3D deep learning,  
researchers predominantly extracted features with 3D-CNN kernels using volumetric representations.
The earlier attempts in this direction could only process voxel-grids of low resolution (e.g.~30$\times$30$\times$30 in ShapeNets~\cite{wu20153d}, 32$\times$32$\times$32 in VoxNet~\cite{maturana2015voxnet}), even with  the modern GPUs.
This issue also transcended to the subsequent works along this direction~\cite{huang2016point,sedaghat2016orientation,zeng20163dmatch,zhang2017deepcontext}. 
The limitation of low input resolution was a natural consequence of the cubic growth of memory and computational requirements associated with the dense volumetric inputs. 
Different solutions later appeared to address these issues. For example,
Engelcke \emph{et al.}~\cite{EngelckeICRA2017} introduced sparsity in the input and hidden neural activations. Their solution is effective in reducing the number of convolutions, but not the amount of required memory. Li \emph{et al.}~\cite{li2016fpnn} proposed a field probing neural network, which transforms 3D data into intermediate representations with a small set of probing filters. Although this network is able to reduce the computational and memory costs of fully connected layers, the probing filters fail to support weight sharing. Later, 
Riegler \emph{et al.}~\cite{riegler2017octnet} proposed the octree-based OctNet, which represents point clouds with a hybrid of shallow grid octrees (depth = 3). Compared to its dense peers, OctNet reduces the computational and memory costs to a large degree, and is applicable to high-resolution inputs up to 256$\times$256$\times$256. However, it still has to perform unnecessary computations in the empty spaces around the objects. 
{\color{\hRevise}SparseConvNet \cite{graham20183d} and MinkowskiNet \cite{choy20194d} exploit sparse tensors to represent the objects with their 
occupied voxels only. By applying the 3D-CNN kernels sparsely to those occupied voxels, they avoid 
unnecessary computations in the empty spaces.
Pointwise CNN \cite{hua2018pointwise} computes the feature of each voxel as the
normalization of features of all points falling into the voxel, which is computationally expensive.} Other recent techniques also transform the original point cloud into other regular representations like tangent image \cite{tatarchenko2018tangent}, {\color{\hRevise}``translating'' tensor \cite{atzmon2018point}} or high-dimensional lattice \cite{su2018splatnet} such that the standard CNNs can be applied to the transformed data. 
{\color{\hRevise}In particular, PCNN \cite{atzmon2018point} extends the features defined on sparse point clouds to functions defined over the entire Euclidean space ($\mathbb{R}^3$).
Its convolution strategy transforms the irregular points into a regular ``translating'' tensor, whose size is quadratic in the size of the point cloud. This leads to a significant computation and memory overhead, making PCNN not scalable.}

\vspace{-2mm}
\subsection{Graph Convolutional Networks}
The demand of irregular data processing with CNN-like architectures has resulted in a recent  rise of graph convolutional networks~\cite{bronstein2017geometric}. 
In general, the broader graph-based deep learning has also seen techniques besides convolutional networks that update vertex features recurrently to propagate the context information (e.g. \cite{gori2005new,scarselli2009graph,li2016gated,qi20173d,ye20183d}). However, here, our focus is on graph convolutional networks that relate to our work more closely.


Graph convolutional networks can be grouped into spectral networks (e.g. \cite{bruna2013spectral,defferrard2016convolutional,kipf2017semi}) and spatial networks (e.g. \cite{simonovsky2017dynamic,wang2018dynamic}). 
The spectral networks perform convolution 
on spectral vertex signals converted from 
Fourier transformation,
while the spatial networks perform convolution directly on the spatial vertices. 
A major limitation of the spectral networks is that they require the graph structure to be fixed, which makes their application to the data with varying graph structures (e.g.~point clouds) challenging. Yi \emph{et al.} \cite{yi2017syncspeccnn} attempted to address this issue with Spectral Transformer Network (SpecTN), similar to STN~\cite{jaderberg2015spatial} in the spatial domain. However, the signal transformation from spatial to spectral domains and vice-versa has computational complexity $\mathcal{O}(n^2)$, resulting in prohibitive requirements for large point clouds.  

ECC~\cite{simonovsky2017dynamic} is among the pioneering works for point cloud analysis with graph convolution in the spatial domain. Inspired by the dynamic filter networks \cite{de2016dynamic}, it adapts MLPs to generate convolution filters between the connected vertices dynamically. 
The dynamic generation of filters naturally comes with a computational overhead. 
{\color{\hRevise}Monte Carlo (MC) convolution \cite{hermosilla2019monte} exploits MLPs as well to learn the edge-wise filters. It otherwise introduces density parameters into the integration which results in a convolution for non-uniformly sampled point cloud processing.
PCCN \cite{wang2018deep} uses MLPs to parameterize its kernel to achieve a continuous convolution. Along similar lines,}
DGCNN \cite{wang2018dynamic}, Flex-Conv \cite{groh2018flex} and SpiderCNN \cite{xu2018spidercnn} explore different parameterizations to generate the edge-dependent filters. Instead of generating filters for the edges individually, few networks also generate a complete local convolution kernel at once using mini networks~\cite{li2018pointcnn, wu2019pointconv}.
Li \textit{et al.}~\cite{li2018pointcnn} recently introduced PointCNN  that uses a  convolution module named $\mathcal{X}$-Conv for point cloud processing. The network achieves good performance on the standard benchmarks (e.g. ShapeNet and S3DIS).
However, the generated kernels are sensitive to the order of neighborhood points indicating that the underlying  representation is not permutation-invariant. Moreover, the strategy of dynamic kernel generation makes the technique computationally inefficient. {\color{\hRevise}Besides the above MLP-based `continuous' convolution kernels, SplineCNN \cite{fey2018splinecnn} also explores to generate the continuous convolution kernel using spline functions.}

{\color{\hRevise}
Discrete kernel is an attractive alternative to the `continuous' kernels in avoiding the computational overhead. In our preliminary work \cite{lei2019octree}, we divide the space in a deterministic and compact manner. In comparison, the recent KPConv kernel \cite{thomas2019kpconv} divides the spherical region into volumetric bins based on template points learned in data independent manner.} 
Wang \emph{et al.}~\cite{wang2019attention} inserted an  attention mechanism in graph convolutional networks to develop GACNet.
Such an extension of graph networks is particularly helpful for semantic segmentation as it enforces the neighborhood vertices to have consistent semantic labels similar to  CRF~\cite{lafferty2001conditional}. 
Besides the convolution operation, graph coarsening and edge construction are two essential parts for the graph network architectures. We briefly review the methods along these aspects below.

\vspace{1mm}
\noindent\textbf{Graph coarsening:} Point cloud sampling methods are useful for graph coarsening. 
PointNet++ \cite{qi2017pointnetplusplus} utilizes farthest point sampling (FPS) to coarsen the point cloud, while 
Flex-Conv~\cite{groh2018flex} samples the point cloud based on inverse densities (IDS) of each point. 
Random sampling is the simplest alternative to FPS and IDS, but it does not perform as well for the challenging tasks like semantic segmentation. 
Recently, researchers also started to explore the possibility of learning sampling with deep neural networks~\cite{dovrat2019learning}. In this work, we exploit FPS as the sampling strategy for graph coarsening, as it does not need training and it reduces the point cloud resolution relatively uniformly.

\vspace{1mm}
\noindent\textbf{Graph connections:} Point neighborhood search can be used to build edge connections in a graph. KNN search generates fixed number of neighborhood points for a given point, which results in a regular graph. Range search generates flexible number of neighborhood points, which may results in irregular graphs. 
Tree structures can also be seen as special kinds of graphs \cite{klokov2017escape,lei2019octree}, however, the default absence of intra-layer connections in trees drastically limits their potential as graph networks.
In a recent example, Rao \emph{et al.}~\cite{rao2019spherical} proposed to employ spherical lattices for regular graph construction. Their technique relies on $1\times 1$ convolution and max-pooling to aggregate the geometric context between neighbouring points.

In this paper, we use range search to establish the graph connections for its natural compatibility with the proposed kernel. Note that our spherical kernel does not restrict the graph vertex degrees to be fixed. Hence, unlike \cite{li2018pointcnn,wu2019pointconv}, our kernel is applicable to both regular and irregular graphs.

\section{Discrete Convolution Kernels} \label{sec:DCK}
Given an arbitrary point cloud of $m$ points  $\mathcal{P}=\{\mathbf{x}_i\in \mathbb{R}^3\}_{i=1}^m$, we
represent the neighborhood of each point $\mathbf{x}_i$ as $\mathcal{N}(\mathbf{x}_i)$. To achieve graph convolution on the target point $\mathbf{x}_i$,
the more common `continuous' filter  approaches~\cite{simonovsky2017dynamic, wang2018dynamic,xu2018spidercnn, groh2018flex,li2018pointcnn,wu2019pointconv} parameterize convolution 
as a function of local point coordinates. 
For instance, suppose $\mathbf{w}$ is the  filter that computes the output feature of  channel $c$. These techniques may represent the filter as $\mathbf{w} =\mathbf{h}(\mathbf{x}_j-\mathbf{x}_i)$, where $\mathbf{h}(.)$ is a continuous function (e.g. MLP) and $\mathbf{x}_j\in\mathcal{N}(\mathbf{x}_i)$. 
However, compared to the continuous filters, discrete kernel is predefined and it does not need the above mentioned (or similar) 
intermediate computations.
This makes a discrete kernel  computationally more attractive. 

{\color{\hRevise}Following the 
standard CNN kernels, a primitive discrete kernel for 
point clouds can be defined on regular grids in the Euclidean space,
similar to the 3D-CNN kernels for voxel-grids \cite{wu20153d,maturana2015voxnet}.}
For resolution $h$, this kernel comprises $h^3$ weight filters $\mathbf{w}_{\kappa\in\{1,\dots,h^3\}}$. By incorporating the notion of separable convolution \cite{chollet2017xception} into this design, each weight filter is transformed  from a vector $\mathbf{w}_\kappa$ to a scalar $w_\kappa$. 
It is noteworthy that the application of a discrete kernel to `graph' representation is significantly different from its volumetric counterpart. Hence, to differentiate, we refer to a kernel for graphs as CNN3D kernel. 
A CNN3D kernel indexes the bins and for the $\kappa^{\text{th}}$ bin, it uses $w_\kappa$ to propagate features from all neighboring point $\mathbf{x}_j, \forall j$ in that bin to the target point $\mathbf{x}_i$, see
Fig.~\ref{fig:discrete_kernel}.
{\color{\hRevise}It performs convolutions based on the existing points in the point cloud. Since no points are populated at empty spaces, the CNN3D kernel can avoid unnecessary computations at
empty spaces. In contrast, the 3D-CNN kernel demands volumetric representations to 
assign an occupancy feature $0$ or $1$ respectively to each voxel in the empty or non-empty spaces around the objects. To fulfill the convolution,
it requires computations at all voxels within the receptive field, whether it is occupied or unoccupied.}

We make the following observation in relation to improving the CNN3D kernels. 
For images, the more primitive constituents, i.e.~patches, have traditionally been used to extract 
hand-crafted features \cite{lowe2004distinctive,dalal2005histograms}. The same principle transcended to the receptive fields of automatic feature extraction with CNNs, which compute feature maps using the activations of well-defined rectangular regions of images. Whereas rectangular regions are intuitive  choice for images, spherical regions are more suited to process unstructured 3D data such as point clouds. Spherical regions are inherently amenable to computing geometrically meaningful features for such data \cite{frome2004recognizing,tombari2010uniqueACM,tombari2010unique}.
Inspired by this natural kinship, we introduce the concept of \emph{spherical convolution kernel}\footnote{The term \emph{spherical} in Spherical CNN~\cite{cohen2018spherical} is used for surfaces (i.e.~$360^\circ$ images) not the ambient 3D space. Our notion of spherical kernel is widely dissimilar, and it is used in a different context. Also, note that, different from the preliminary work~\cite{lei2019octree}, here the spherical kernel is only used to perform depth-wise spatial convolutions.} (termed SPH3D kernel) that considers a 3D sphere as the basic geometric shape to perform the convolution operation. 
We explain the proposed discrete spherical kernel in Section~\ref{subsec:SPH3D}, and later contrast it to the existing CNN3D kernels in Section~\ref{subsec:SPH3D_CNN3D}.

\subsection{Spherical Convolutions} \label{subsec:SPH3D}
\begin{figure*}[!t]
  \centering
  \includegraphics[width=0.96\textwidth]{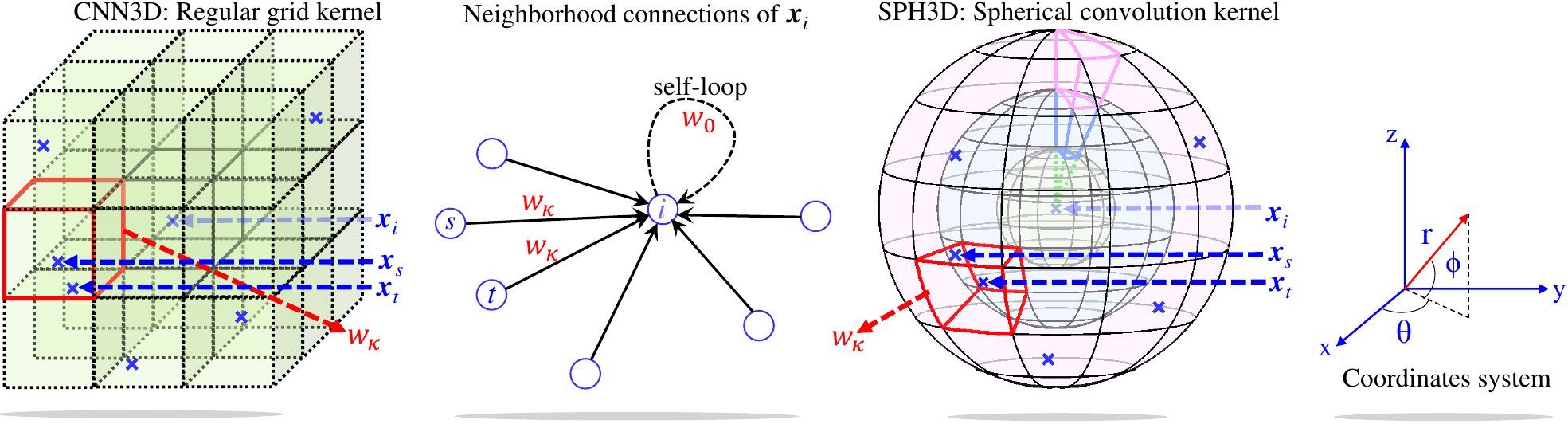}
  \vspace{-3mm}
  \caption{Illustration of the primitive CNN3D and the proposed SPH3D discrete kernels: The point $\mathbf{x}_i$ has  seven neighboring points including itself (the self-loop). To perform convolution at $\mathbf{x}_i$, discrete kernels systematically partition the space around it into bins. With $\mathbf{x}_i$ at the center, CNN3D divides a 3D `cubic' space around the point into `uniform voxel' bins. Our SPH3D kernel partitions a `spherical' space around  $\mathbf{x}_i$ into `non-uniform volumetric' bins.
  For both kernels, the bins and their corresponding weights are indexed. The points falling in the $\kappa^{\text{th}}$ bin are propagated to $\mathbf{x}_i$ with the weight $w_\kappa$. Multiple points falling in the same bin, e.g.~$\mathbf{x}_s$ and $\mathbf{x}_t$, use the same weight for computing the output   feature at $\mathbf{x}_i$.}
  \label{fig:discrete_kernel}
  \vspace{-2mm}
\end{figure*}

We define the convolution kernel with the help of a sphere of radius $\rho\in \mathbb{R}^+$, see Fig.~\ref{fig:discrete_kernel}. For a target point $\mathbf{x}_i$, we consider its neighborhood $\mathcal{N}(\mathbf{x}_i)$ to be the set of points within the sphere centered at $\mathbf{x}_i$,  i.e. $\mathcal{N}(\mathbf{x}_i)=\{\mathbf{x}:d(\mathbf{x},\mathbf{x}_i)\leq \rho\}$, where $d(.,.)$ is a distance metric - $\ell_2$ distance in this work.
We divide the sphere into $n \times p \times q$ \emph{`bins'} by partitioning the occupied space uniformly along the azimuth ($\theta$) and elevation ($\phi$) dimensions. We allow the partitions along the radial ($r$) dimension to be non-uniform because the cubic volume growth for large radius values can be undesirable.
Our quantization of the spherical region is mainly inspired by 3DSC~\cite{frome2004recognizing}.
We also define an additional bin corresponding to the origin of the sphere to allow the case of self-convolution of points on the graph.
To produce an output feature map, we define a learnable weight parameter $w_{\kappa \in \{0, 1,\dots,n \times p \times q \}}\in\mathbb{R}$ for  each bin, where $w_{0}$ relates to self-convolution.  
Combined, the $n \times p \times q + 1$ weight values specify a single spherical convolution kernel. 

To compute the activation value for a target point $\mathbf{x}_i$, we first identify the relevant weight values of its neighboring points $\mathbf{x}_j\in \mathcal{N}(\mathbf{x}_i)$.
It is straightforward to associate $w_0$ to $\mathbf{x}_i$ for self-convolution. For the non-trivial cases, we first represent the neighboring points in terms of their spherical coordinates that are referenced using $\mathbf{x}_i$ as the origin. That is, for each $\mathbf{x}_j$ we compute $\mathcal T(\boldsymbol{\Delta}_{ji})\rightarrow \boldsymbol{\psi}_{ji}$, where $\mathcal T(.)$ defines the transformation from Cartesian to Spherical coordinates and  $\boldsymbol{\Delta}_{ji}=\mathbf{x}_j-\mathbf{x}_i$.
Supposing that the bins of the quantized sphere are respectively indexed by $k_\theta$, $k_\phi$ and $k_r$ along the azimuth, elevation and radial dimensions, the weight values associated with each spherical kernel bin can then be indexed as $\kappa = k_\theta + (k_\phi-1)\times n + (k_r-1)\times n\times p$, where $k_\theta \in \{1,\dots,n\},~k_\phi \in \{1,\dots,p\},~k_r \in \{1,\dots,q\}$. Using this indexing, we relate the relevant weight value to each $\boldsymbol{\psi}_{ji}$, and hence $\mathbf{x}_j$.
In the $l^{\text{th}}$ network layer, the activation for the $i^{\text{th}}$ point in channel $c$ gets computed as:
\begin{align}\label{equ:spatio-conv}
  &z^l_{ic}  = \frac{1}{|\mathcal{N}(\mathbf{x}_i)|}\sum\limits_{j = 1}^{|\mathcal{N}(\mathbf{x}_i)|}w^l_{\kappa c} a^{l-1}_{jc}+b_{c},\\
  &a^l_{ic} = f(z^l_{ic}),
  \end{align}
where $a^{l-1}_{jc}$ is the feature of a neighboring point from layer $l-1$, $w^l_{\kappa c}$ is the weight value, and $f(\cdot)$ is the non-linear activation function - ELU\cite{clevert2016fast} in our experiments. 
By applying the spherical convolution $\lambda$ times for each input channel, we produce $\lambda C_{in}$ output features for the target convolution point $\mathbf{x}_i$.  {\color{\hRevise}We note that based on the geometry of points, different number of points may fall in different bins of our kernel. The  distinctive influence of points resulting from this distribution encodes the local structure of data, which contributes towards learning discriminative features under the overall learning objective of our network.}

To elaborate on the characteristics of the spherical convolution kernel, we denote the \emph{boundaries} along $\theta$, $\phi$ and $r$ dimensions of the kernel bins as follows:
\begin{align}
\notag
&\boldsymbol{\Theta}=[\Theta_1,\dots,\Theta_{n+1}], ~\Theta_k<\Theta_{k+1}, {\Theta}_k\in[-\pi,\pi], \\
\notag
&\boldsymbol{\Phi}=[\Phi_1,\dots, \Phi_{p+1}]\big],~\Phi_k<\Phi_{k+1}, {\Phi}_k\in\big[-\frac{\pi}{2}, \frac{\pi}{2}] \\
\notag
&\mathbf{R}=[R_1,\dots,R_{q+1}], ~~R_k<R_{k+1}, R_k\in(0,\rho].
\end{align}
The constraint of uniform splitting along the azimuth and elevation
results in ${\Theta}_{k+1}-{\Theta}_k=\frac{2\pi}{n}$ and ${\Phi}_{k+1}-{\Phi}_k=\frac{\pi}{p}$.
\vspace{1mm}
{\color{\hRevise}\noindent {\bf Lemma 2.1:} \emph{If~$\Theta_k\cdot\Theta_{k+1}\geq 0$,  $\Phi_k\cdot\Phi_{k+1}\geq 0$ and $n>2$, then for any two neighboring points $\mathbf{x}_a\neq\mathbf{x}_b$, where the spherical convolution kernel is centered at either $\mathbf{x}_a$ or $\mathbf{x}_b$, the weight value $w_{\kappa}, \forall \kappa>0$, are applied asymmetrically}.}

\vspace{1mm}
\noindent{\it Proof:}
Let $\boldsymbol{\Delta}_{ab}= {\bf x}_a - {\bf x}_b =[\delta_x,\delta_y,\delta_z]^{\intercal}$, then $\boldsymbol{\Delta}_{ba}=[-\delta_x,-\delta_y,-\delta_z]^{\intercal}$.  Under the Cartesian to Spherical coordinate transformation,  we have $\mathcal T(\boldsymbol{\Delta}_{ab}) =\boldsymbol{\psi}_{ab}=[\theta_{ab},\phi_{ab},r]^{\intercal}$,  and $\mathcal T(\boldsymbol{\Delta}_{ba}) = \boldsymbol{\psi}_{ba}=[\theta_{ba},\phi_{ba},r]^{\intercal}$. Assume that the resulting $\boldsymbol{\psi}_{ab}$ and $\boldsymbol{\psi}_{ba}$ fall in the same bin indexed by $\kappa \leftarrow (k_\theta,k_\phi,k_r)$, i.e. $w_{\kappa}$ will have to be applied symmetrically to the original points.
In that case, under the inverse transformation $\mathcal T^{-1} (.)$, we have $\delta_z=r\sin\phi_{ab}$ and $(-\delta_z)=r\sin\phi_{ba}$.
The condition $\Phi_{k_\phi}\cdot\Phi_{k_\phi+1}\geq 0$ entails that $-\delta_z^2 = \delta_z\cdot(-\delta_z)=(r\sin\phi_{ab})\cdot(r\sin\phi_{ba})=r^2(\sin\phi_{ab}\sin\phi_{ba})\geq 0\Longrightarrow\delta_z=0$. Similarly,
$\Theta_{k_\theta}\cdot\Theta_{k_\theta+1}\geq 0 \Longrightarrow \delta_y=0$.
Since $\mathbf{x}_a\neq\mathbf{x}_b$, for  $\delta_x\neq0$ we have $ \cos\theta_{ab} = -\cos\theta_{ba} \Longrightarrow |\theta_{ab}-\theta_{ba}|=\pi$.
However, if  $\theta_{ab}$, $\theta_{ba}$ fall into the same bin, we have $|\theta_{ab}-\theta_{ba}|=\frac{2\pi}{n}<\pi$, which entails $\delta_x = 0$.  Thus, $w_{\kappa}$ can not be applied to any two points symmetrically  unless both points are the same.
\vspace{1mm}

{\color{\hRevise}The kernel asymmetry forbids weight sharing between point pairs for the convolution operation, which leads to learning fine geometric details of the point clouds. In fact, asymmetry exists widely in the standard kernels, e.g.~2D-CNN, 3D-CNN for voxel-grids. However, for the discrete SPH3D kernel, not all partitionings guarantee asymmetry. Lemma~2.1 provides guidelines on how to divide the spherical space into kernel bins such that the asymmetry is always preserved.
The importance of kernel asymmetry is also demonstrated quantitatively in Section~\ref{subsec:SPH3D_CNN3D}.}
The resulting division also ensures  translation-invariance of the kernel, similar to the standard CNN kernels. 
Additionally, unlike the convolution operation of PointCNN~\cite{li2018pointcnn}, the proposed kernel is invariant to point permutations because it explicitly incorporates the geometric relationships between the point pairs.


We can apply the spherical convolution kernel to learn depth-wise features in the spatial domain. The point-wise convolution can be 
readily achieved with shared MLP or $1\times 1$ convolution 
using any modern deep learning library.
To be more precise, the two convolutions make our kernel perform \emph{separable} convolution \cite{chollet2017xception}. However, we generally refer to it as spherical convolution,  for simplicity.

\begin{table*}[ht]
    \centering
    \caption{{\color{\hRevise}Segmentation results on Area 5 of S3DIS with different CNN3D kernels, their variants with self-convolution weight $w_0$, and our SPH3D kernel. Input point cloud size 2048 is used.}}
    \label{tab:CNN3D_vs_SPH3D}
    \begin{adjustbox}{width=0.8\textwidth}
    \begin{tabular}{l|c|c|c|c|c|c}
    \hline
       Kernel type & \multicolumn{3}{c|}{CNN3D}& \multicolumn{2}{c|}{CNN3D + $w_0$}  & SPH3D\\
       \hline
        Networks & 
       Network-1 &  Network-2 & Network-3&  Network-4 & Network-5&Network-6\\
       \hline
       Kernel size & 
       $3\times3\times3$ & $4\times4\times4$ & $5\times5\times5$ &$3\times3\times3+1$ & $5\times5\times5+1$ &$8\times2\times2+1$\\
       \hline
       \# Bins &27& 64& 125& 28 & 126& 33 \\
       \hline
        OA & 85.7& 87.4& 85.9& 85.4& 85.9 & 87.9\\
        \hline
         mAcc & 65.3 &68.5& 66.9&67.8 & 66.2 & 67.8\\
         \hline
       mIoU & 57.9 & 60.8&58.6& 58.3&58.3  & 61.0\\
       \hline
    \end{tabular}
    \end{adjustbox}
    \vspace{-2mm}
\end{table*}

\subsection{Comparison to CNN3D kernel} \label{subsec:SPH3D_CNN3D}
CNN3D kernel rasterizes 3D data into uniform voxel grids, where  the size of 
$3\times3\times3=27$ is prevalently used.  
This size splits the space in 1 voxel for radius $\rho=0$ (self-convolution); 6 voxels for radius $\rho=1$;
  12 voxels for radius $\rho=\sqrt{2}$; and 8 voxels for radius $\rho=\sqrt{3}$.
An analogous spherical convolution kernel for the same region can be specified with a radius $\rho=\sqrt{3}$, using the following edges for the bins:
 \begin{align}\label{split-sphConv}
 \notag
 &\boldsymbol{\Theta}=[-\pi,-\frac{\pi}{2},0,\frac{\pi}{2},\pi];\\
 \notag
&\boldsymbol{\Phi} =[-\frac{\pi}{2},-\frac{\pi}{4},0,\frac{\pi}{4},\frac{\pi}{2}];\\
&\mathbf{R} = [\epsilon,1,\sqrt{2},\rho], \epsilon\rightarrow0^+.
  \end{align}
  This division results in a \emph{kernel size} (i.e. total number of bins) $4\times 4\times 3 + 1=49$, which is one of the coarsest multi-scale quantization allowed by Lemma~2.1. 

  Notice that, if we move radially from the center to periphery of the spherical kernel, we encounter identical number of bins (16 in this case) after each edge defined by $\mathbf{R}$, where fine-grained bins are located close to the origin that can encode detailed local geometric information of the points. This is in sharp contrast to CNN3D kernels that must keep the size of all cells constant and rely on increased resolution to capture the finer details. This makes their number of parameters grow cubicly, harming the scalability.
The multi-scale granularity of spherical kernel (SPH3D) allows for more compact  representation. 


{\color{\hRevise}
To corroborate, we briefly touch upon semantic segmentation with CNN3D and SPH3D kernel, using a popular benchmark dataset S3DIS \cite{armeni20163d} in Table~\ref{tab:CNN3D_vs_SPH3D}. 
We give further details on the dataset and experimental settings in Section~\ref{sec:Exp}. 
Here, we focus on the aspect of representation compactness resulting from the non-uniform granularity of the bins in SPH3D. 
In the table, the only difference in the networks is in the used kernels. All the other experimental details are `exactly' the same for all networks. 
Network-1 and 3 use CNN3D kernels that partition the space into $3\times 3\times 3=27$ and $5\times 5\times 5=125$ bins, respectively.
Network-4 and 5 use similar CNN3D kernels but with an additional self-convolution weight $w_0$, resulting in  kernels of $3\times 3\times 3+1=28$ and $5\times 5\times 5+1=126$ bins, respectively. Comparing the results of Network-1 to 4 and Network-3 to 5, we notice that self-convolution weight $w_0$ has minor influence on the final performance of CNN3D kernels.
The SPH3D kernel partitions the space into $8\times 2\times 2 + 1=33$ bins.
Consequently, the kernel requires $1.22\times$ and $1.18\times$ parameters as compared to the Network-1 and 4 kernels, but only $0.26\times$ parameters required by the Network-3 and 5 kernels. 
However, the performance of Network-6 is much better than Network-1 and 3, and also their variants with self-convolution weights. Such an advantage is a natural consequence of the non-uniform partitioning allowed by our kernel.
}

{\color{\hRevise}We compare the representation compactness of SPH3D kernel to odd-size CNN3D kernels in the above experiments. Similar to the proposed SPH3D kernel, not all partitionings of the primitive CNN3D kernel guarantee asymmetry. 
In particular, the \emph{even}-size CNN3D kernels preserve asymmetry while the \emph{odd}-size kernels are unable to do so.
We show the importance of asymmetry by exploring CNN3D kernels of sizes $4\times 4\times 4$ and $5\times 5\times 5$. 
Comparing the performance of Network-2 and 3 in Table~1, we notice that although $4\times 4\times 4$ kernel uses less parameters than the $5\times 5\times 5$ kernel, it still produces better results. This is because the $4\times 4\times 4$ kernel preserves asymmetry and hence learns more representative features. 
Lemma 2.1 ensures that our SPH3D kernel preserves asymmetry as well. The asymmetry, together with the non-uniform multi-scale granularity, results in Network-6 to easily match the performance of Network-2 with only 0.52$\times$ parameters.}

\begin{figure*}[!t]
  \centering
  \includegraphics[width=0.96\textwidth]{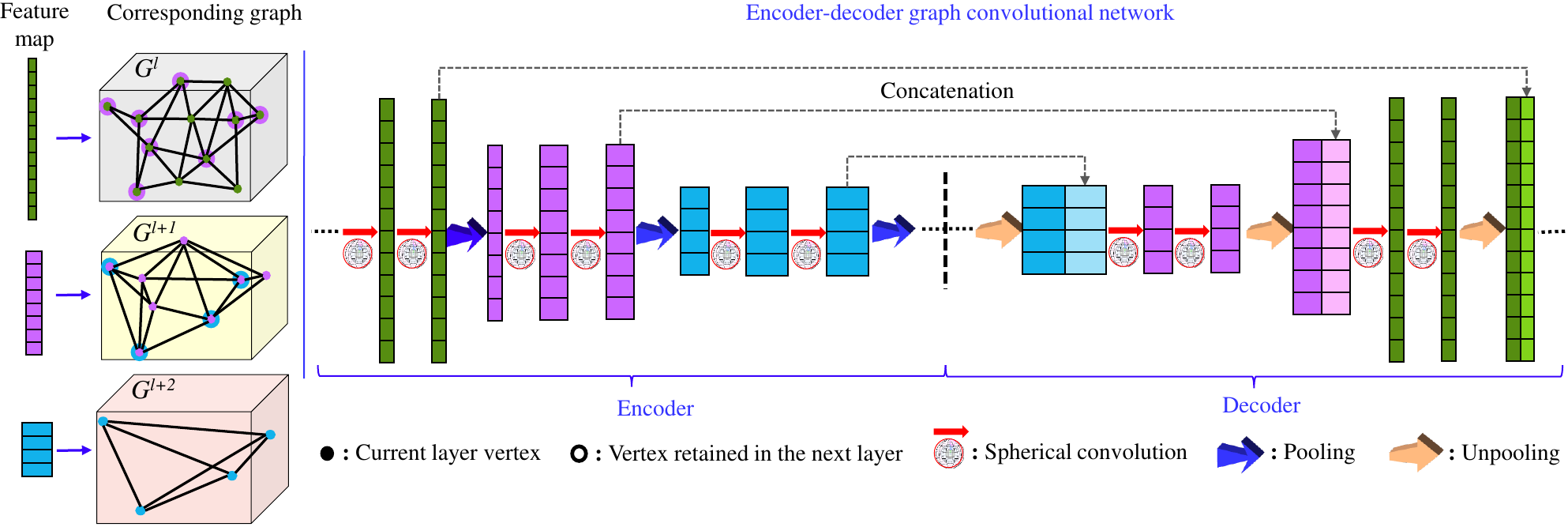}
  \caption{Illustration of Encoder-decoder graph neural network for a toy example. A graph $G^l$ of 12 vertices gets coarsened to $G^{l+1}$ (8 vertices) and further to $G^{l+2}$ (4 vertices),  and expanded back to 12 vertices. The width variation of feature maps depicts different number of feature channels, whereas the number of cells  indicates the total vertices in the corresponding graph. The pooling/unpooling  operations compute features of the coarsened/expanded  graphs. Consecutive  convolutions are applied to form convolution blocks. The shown architecture for semantic segmentation uses skip connections for feature concatenation, similar to U-Net. For classification, the decoder and skip connections are removed and a global 
  representation is fed to a classifier. We omit self loops in the shown graphs for clarity.   
}
  \label{fig:graph_arch}
\end{figure*}

\section{Graph Neural Network}\label{sec:graphNN}
In this work, we employ graph neural network to process point clouds. Compared to the inter-layer connectivity  of the octree-guided network of our  preliminary work~\cite{lei2019octree}, graph representation additionally allows for intra-layer connections. This is beneficial in defining effective convolutional blocks as well as pooling/unpooling modules in the network.   
Let us consider 
a graph $G=(V,E)$ constructed
from a point cloud $\mathcal{P}=\{\mathbf{x}_1,\dots,\mathbf{x}_m\}$, where $V=\{1,2,\dots,m\}$ and $E\subseteq |V|\times|V|$ respectively represent the sets of vertices and edges. It is straightforward to associate each vertex $i\in V$ of the graph  to a point location $\mathbf{x}_i$ and its corresponding feature $\mathbf{a}_i$. However, the edge set $E$ must be carefully established based on the neighborhood of the points. 

\vspace{1mm}
\noindent\textbf{Edge construction:} 
We use range search with a specified radius $\rho$ to get the spatial neighborhood of each point and construct the edge connections of each graph vertex. 
In the range search, neighborhood computations are independent of each other, which makes the search suitable for parallel processing and taking advantage of GPUs. 
{\color{\hRevise}The time complexity of constructing neighborhoods of a single point is linear in the number of vertices $|V|$, which makes  $\mathcal{O}(|V|^2)$ to be the time complexity for a point cloud. {\color{\hReviseMinor}However, the parallelism of modern GPU 
makes the neighbor search efficient enough for reasonably large input point clouds.} 
We provide the neighbour search time for varying point cloud sizes as the `graph construction' time in Table 10 (Section 7).} 
One potential problem of using range search is that large number of neighborhood points in dense clouds can cause memory issues.
We sidestep this problem by restricting the number of neighboring points to $K\in \mathbb{Z}^+$ by randomly sub-sampling the neighborhood, if required. 
The edges are finally built on the sampled points. 
As a result, the neighborhood indices of the $i^{\text{th}}$ vertex can be denoted as $\mathcal{N}(i)=\{j:(j,i)\in E\}$, in which $|\mathcal{N}(i)|\leq K$.
With these  sets identified, we
can later compute features for vertices with spherical convolution. 

\vspace{1mm}
\noindent\textbf{Graph coarsening:} 
We use Farthest Point Sampling (FPS) to coarsen the point graph in our network layer-by-layer.
The FPS algorithm  selects one random seed vertex, and iteratively searches for the point that is farthest apart from the previously selected points for the sampling purpose. The algorithm terminates when the desired number of sampled points are acquired, which form the coarsened graph.
By alternately constructing the edges and coarsening the graph for $l_{\max}$ times, we construct a graph pyramid composed of $l_{\max}+1$ graphs, i.e. $G^0 \rightarrow G^1\rightarrow \dots\rightarrow G^{l_{\max}-1}\rightarrow G^{l_{\max}}$.
As compared to the octree structure based graph coarsening adopted in the preliminary work~\cite{lei2019octree}, FPS coarsening has the advantage of keeping the number of vertices of each layer fixed across different samples, which is conducive for more systematic application of convolutional kernels. 
\vspace{1mm}
\noindent\textbf{Pooling:} 
Once a graph is coarsened, we still need to compute the features associated with its vertices. To that end, we define max pooling and average pooling operations to sample features for the coarsened graph vertices. 
Inter-layer graph connections facilitate these  operations.
To be consistent, we denote the graphs before and after pooling layer $l\in\{1,\dots,l_{\max} \}$ as $G^{l-1}=(V^{l-1},E^{l-1})$ and $G^l=(V^l,E^l)$ respectively, where $V^{l-1}\supset V^l$. Let $i^{l-1}\in V^{l-1}$ and $i^l\in V^l$ be the two vertices associated with the same point location. The  inter-layer neighborhood of $i^l$ can be readily constructed from graph $G^{l-1}$ as  $\mathcal{N}(i^l)=\{j:(j,i^{l-1})\in E^{l-1}\}$. We denote the features of $i^l$ and its neighborhood point $j\in \mathcal{N}(i^l)$ as
$\mathbf{a}_i^l=[\dots,a_{ic}^l,\dots]^{\intercal}$ and  $\mathbf{a}_j^{l-1}=[\dots,a_{jc}^{l-1},\dots]^{\intercal}$ respectively. The max pooling operation then computes the feature of the vertex $i^l$ as 
\begin{equation}\label{equ:max}
a_{ic}^l = \max\{a_{jc}^{l-1}:j\in\mathcal{N}(i^l)\},  
\end{equation}
while the average pooling computes it as 
\begin{equation}\label{equ:average}
a_{ic}^l =\frac{1}{|\mathcal{N}(i^l)|} \sum\limits_{j\in\mathcal{N}(i^l)} a_{jc}^{l-1}.
\end{equation}
We introduce both pooling operations in our source code release, but use max pooling in our experiments as it is commonly known to have superior performance in point cloud processing~\cite{qi2017pointnet, qi2017pointnetplusplus,shen2018mining}.

\vspace{1mm}
\noindent\textbf{Unpooling:} Decoder architectures with increasing neuron resolution are important for element-wise predictions in semantic segmentation~\cite{badrinarayanan2017segnet}, dense optical flow~\cite{dosovitskiy2015flownet}, etc. 
We build graph decoder by inverting the graph pyramid as  $G^{l_{\max}-1}\rightarrow \dots \rightarrow G^1\rightarrow G^0$. The coarsest graph $G^{l_{\max}}$ is ommited in the reversed pyramid because it is shared between encoder and decoder. 
We denote the graphs before and after an unpooling layer $l\in \{1,\dots,l_{\max} \}$ as
$G^{l_{\max}-l+1}$ and $G^{l_{\max} -l}$ respectively.
To upsample the features from  $G^{l_{\max}-l+1}$ to  $G^{l_{\max}-l}$, 
we define two types of feature interpolation operations, namely; uniform interpolation and weighted interpolation. 
Notice that the neighborhood set $\mathcal{N}(.)$ in Eqs. (\ref{equ:max}), (\ref{equ:average}) is readily available because of the relation $V^{l-1}\supset V^l$.
However, 
the vertices of graphs $G^{l_{\max}-l+1}$ and $G^{l_{\max}-l}$ satisfy $V^{l_{\max}-l+1}\subset V^{l_{\max}-l}$ on the contrary.
Therefore, we have to additionally construct the neighborhood  of $i^{l_{\max}-l}\in V^{l_{\max}-l}$ from $V^{l_{\max}-l+1}$. For that, we again use the range search to compute $\mathcal{N}(i^{l_{\max}-l})$.
The features of $i^{l_{\max}-l}$ and its neighborhood points  $j\in\mathcal{N}(i^{l_{\max}-l})$ can be consistently denoted as 
$a_{ic}^{l_{\max}+l}$ and $a_{jc}^{l_{\max}+l-1}$.
The \textit{uniform} interpolation computes the feature of vertex $i^{l_{\max}-l}$ as the average features of its inter-layer neighborhood points, i.e.
\begin{equation}\label{eq:uniform_interpolate}
a_{ic}^{l_{\max}+l} = \frac{1}{|\mathcal{N}(i^{l_{\max}-l})|} \sum \limits_{j\in\mathcal{N}(i^{l_{\max}-l})} a^{l_{\max}+l-1}_{jc}.
\end{equation}
The \textit{weighted} interpolation computes the features of vertex $i^{l_{\max}-l}$  by weighing its
neighborhood features based on
their distance to $i^{l_{\max}-l}$. Mathematically, 
\begin{equation}\label{eq:weighted_interpolate}
a_{ic}^{l_{\max}+l} =
\frac{\sum\limits_{j\in\mathcal{N}(i^{l_{\max}-l})}w_{ji}a^{l_{\max}+l-1}_{jc}}{\sum\limits_{j\in\mathcal{N}(i^{l_{\max}-l})} w_{ji}},
\end{equation}
where $w_{ji}=d(\mathbf{x}_j^{l_{\max}-l+1},\mathbf{x}_i^{l_{\max}-l})$. Here, {\color{\hRevise}$d(.,.)$ is the $\ell_2$ distance function} and the points $\mathbf{x}_j^{l_{\max}-l+1}$ and 
$\mathbf{x}_i^{l_{\max}-l}$ are associated to vertices $j$ and  $i^{l_{\max}-l}$, respectively. In our source code, we provide both types of interpolation functionalities for upsampling. However, the experiments in Section~\ref{sec:Exp} are performed with uniform interpolation for its computational efficiency.



In Fig.~\ref{fig:graph_arch}, we illustrate an encoder-decoder graph neural network constructed by our technique for a \textit{toy} example.
In the shown network, a graph $G^l$ of 12 vertices gets coarsened to 8 ($G^{l+1}$) and 4 ($G^{l+2}$) vertices in the encoder network, and later gets  expanded in the decoder network.  
The pooling/unpooling operations are applied to learn features of the structure altered graphs.  
The graph structure remains unchanged during convolution operation. Notice, we apply consecutive spherical convolutions to form convolution \textit{blocks} in our networks. 
In the figure, variation in width of the feature maps depicts different number of channels (e.g. 128, 256 and 384) for the features.
The shown U-shape architecture for the task of semantic segmentation also exploits skipping connections similar to U-Net~\cite{ronneberger2015u,badrinarayanan2017segnet}. These connections copy features from the encoder and concatenate them to the decoder features. For the classification task, these connections and the decoder part are removed, and a global feature representation is fed to a classifier comprising fully connected layers. 
The simple architecture in  Fig.~\ref{fig:graph_arch} graphically illustrates the application of the above-mentioned concepts to our networks in Section~\ref{sec:Exp}, where we provide  details of the architectures used in our experiments. 


\vspace{1mm}
\noindent\textbf{Software for Tensorflow:}
With this article, we also release a cuda enabled implementation for the above presented concepts. The package is  Tensorflow compatible~\cite{abadi2016tensorflow}.
As compared to the Matconvnet~\cite{vedaldi2015matconvnet} source code of the preliminary work~\cite{lei2019octree}, Tensorflow compatibility is chosen due to the popularity of the programming framework.  
In the package, we provide cuda implementations of the spherical convolution, range search, max pooling, average pooling, uniform interpolation and weighted interpolation. 
The provided spherical kernel implementation can be used for convolutions on both regular and irregular graphs. {\color{\hRevise}Unlike the existing methods~(e.g.~\cite{li2018pointcnn,wu2019pointconv}), we do not impose any constraint on the vertex degree of the graph except the upper pragmatic limit $K$, allowing the graphs to be more flexible, similar to ECC
\cite{simonovsky2017dynamic}.}
In our implementation, the spherical convolutions are all followed by batch normalization~\cite{ioffe2015batch}.
In the preliminary work~\cite{lei2019octree}, the implemented spherical convolution does not separate the depth-wise convolution from the point-wise convolution \cite{chollet2017xception}, thereby performing the two convolutions simultaneously similar to a typical convolution operation.
Additionally, the previous implementation is specialized to octree structures, and hence not applicable to general graph architectures. The newly released implementation for Tensorflow improves on all of these aspects. The source code and further details of the released package can be found at \href{https://github.com/hlei-ziyan/SPH3D-GCN}{https://github.com/hlei-ziyan/SPH3D-GCN}.

\begin{figure*}
    \centering
    \includegraphics[width=0.8\textwidth]{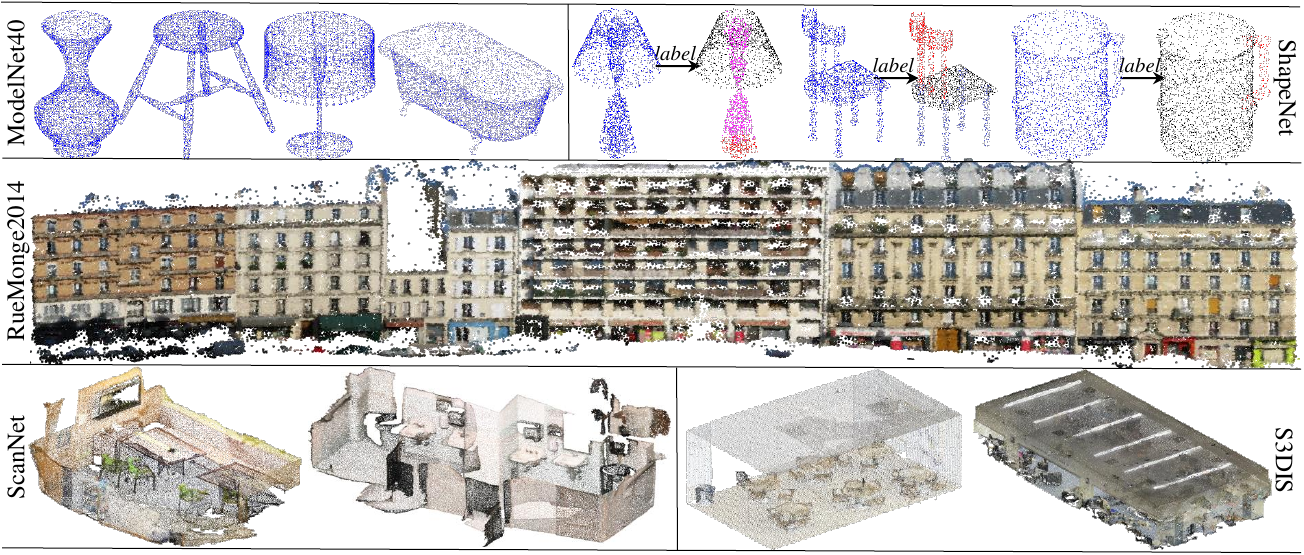}
    \caption{Representative samples from datasets: ModelNet40 and ShapeNet provide point clouds of synthetic models. We also illustrate ground truth segmentation for ShapeNet. 
    RueMonge2014 comprises point clouds for outdoor scenes, while ScanNet and S3DIS contain indoor scenes.   
    }
    \label{fig:datasets}
\end{figure*}

\section{Experiments}
\label{sec:Exp}
We evaluate our technique for classification and semantic segmentation tasks using clean CAD point clouds and  large-scale noisy point clouds of real-world scenes.  
The dataset used in our experiments include ModelNet~\cite{wu20153d}, ShapeNet~\cite{yi2016scalable}, RueMonge2014~\cite{riemenschneider2014learning}, ScanNet~\cite{dai2017scannet} and S3DIS~\cite{armeni20163d}, for which representative samples are illustrated in  
Fig.~\ref{fig:datasets}.
We only use the $(x,y,z)$ coordinates of points to train our networks, except when the $(r,g,b)$ values are also available. In that case, we additionally use those values by rescaling them into the range $[-1,1]$. 
We note that, a few existing methods also take advantage of normals as input features \cite{qi2017pointnet,qi2017pointnetplusplus,li2018so,rao2019spherical}.
However, normals are not directly sensed by the 3D sensors and must be computed separately, entailing additional computational burden. Hence, we avoid using normals as input features except for RueMonge2014, which already provides the normals. 

Throughout the experiments, we apply the spherical convolution  with a  kernel size $8\times2\times2+1${\color{\hRevise}, where the radial dimension is split uniformly}. 
Our network training is conducted on a single Titan Xp GPU with 12 GB memory.
We use Adam Optimizer \cite{kingma2015adam} with an initial learning rate of 0.001 and momentum 0.9 to train the network. The batch size is kept fixed to 32 in ModelNet and ShapeNet, and 16 the remaining datasets.
The maximum neighborhood connections for each vertex is set to $K=64$. These hyper-parameters are empirically optimized with  cross-validation.
We also employ data augmentation in our experiments.
For that, we use random sub-sampling to drop points, and random rotation, which include azimuth rotation (up to $2\pi$ rad) and small arbitrary perturbations (up to $10^\circ$ degrees) to change the view of point clouds. We also apply random scaling, shifting and noisy translation of points with std.~dev = 0.01. 
These operations are commonly found in the related literature. We apply them on-the-fly in each training epoch of the network. 

\begin{table*}[!t]
    \centering
    \caption{Network configuration details: NN($\rho$) denotes a range search with radius $\rho$. {\color{\hRevise}SPH3D($C_{in}$, $C_{out}$, $\lambda$) represents a separable spherical convolution that takes $C_{in}$ input features, performs a depth-wise convolution with a multiplier $\lambda$ followed by a point-wise convolution to generate $C_{out}$ features. When $\lambda$ is omitted in the table, we use  $\lambda=2$. MLP($C_{in}$, $C_{out}$) and FC($C_{in}$, $C_{out}$) indicate multilayer perceptron and fully connected layer taking $C_{in}$ input features, and output $C_{out}$ features.} G-SPH3D denotes global spherical convolution that applies SPH3D once to a single point for global feature learning. The brackets [~] are used to show feature concatenation. The pool($A,B$) and unpool($A,B$) operations transform vertices $A$ into $B$, and $C$ indicates the number of classes in a dataset.}\label{tab:net_config}
    \begin{adjustbox}{width=1\textwidth}
    {\Huge\begin{tabular}{l|c|c|c|c|c|c|c|c|c|c}
    \hline
    Layer Name & MLP1 &  Encoder1 &Encoder2&Encoder3&Encoder4&Decoder4&Decoder3    &Decoder2&Decoder1 &Output\\
    \hline
       \multirow{4}{*}{ModelNet40}  & &  NN($\rho=0.1$) &NN($\rho=0.2$)&NN($\rho=0.4$)&&\multirow{4}{*}{--}&\multirow{4}{*}{--}&\multirow{4}{*}{--}&\multirow{4}{*}{--}&FC(832,512) \\
       &MLP  &  SPH3D(64,64) &SPH3D(64,64,1)&SPH3D(128,128,1)&G-SPH3D&  && &&FC(512,256)\\
      & (3,32) &  SPH3D(64,64,1) &SPH3D(64,128)&SPH3D(128,128,1)&(128,512)&&& &&FC(256,40)\\
      &  & pool(10K,2500)&pool(2500,625)&pool(625,156)&&&&&&\\
       \hline
       \multirow{4}{*}{ShapeNet}  &  &  NN($\rho=0.08$) &NN($\rho=0.16$)&NN($\rho=0.36$)&NN($\rho=0.64$)&&\big[Enc4,Dec4\big]&\big[Enc3,Dec3\big] &\big[Enc2,Dec2\big]&\big[Enc1,Dec1\big] \\
       & MLP &  SPH3D(64,128) &SPH3D(128,256)&SPH3D(256,256)&SPH3D(256,512)&SPH3D(512,512)&SPH3D(1024,256)&SPH3D(512,256)&SPH3D(512,128)  &[MLP1,MLP\\
      &(3,64)  &  SPH3D(128,128) &SPH3D(256,256)&SPH3D(256,256)&SPH3D(512,512)&SPH3D(512,512)&SPH3D(256,256)&SPH3D(256,256)&SPH3D(128,128)&(256,64)]\\
      &  & pool(2048,1024) &pool(1024,768)&pool(768,384)&pool(384,128)&unpool(128,384)&unpool(384,768)    &unpool(768,1024) &unpool(1024,2048)&FC(128,$C$)\\
       \hline
       RueMonge- &MLP
       &  NN($\rho=0.1$) &NN($\rho=0.2$)&NN($\rho=0.4$)&NN($\rho=0.8$)&&\big[Enc4,Dec4\big]&\big[Enc3,Dec3\big]   &\big[Enc2,Dec2\big]&\big[Enc1,Dec1\big] \\
        2014& (9,64) &  SPH3D(64,128) &SPH3D(128,256)&SPH3D(256,256)&SPH3D(256,512)&SPH3D(512,512)&SPH3D(1024,256)  &SPH3D(512,256)&SPH3D(512,128) &FC(256,$C$)\\
        \cline{1-2}
      ScanNet, & MLP  &  SPH3D(128,128) &SPH3D(256,256)&SPH3D(256,256)&SPH3D(512,512)&SPH3D(512,512)&SPH3D(256,256)&SPH3D(256,256)&SPH3D(128,128) &\\
      S3DIS& (6,64) & pool(8192,2048) &pool(2048,768)&pool(768,384)&pool(384,128)&unpool(128,384)&unpool(384,768)    &unpool(768,2048)&unpool(2048,8192) &\\
       \hline
    \end{tabular}}
    \end{adjustbox}
\end{table*}

\vspace{1mm}
\noindent\textbf{Network Configuration:}
Table \ref{tab:net_config} provides the summary of network configurations used in our experiments for the classification and segmentation tasks. 
We use identical configurations for semantic segmentation on the realistic datasets RueMonge2014, ScanNet and S3DIS, but a different one for the part segmentation of the synthetic ShapeNet. Our network for the realistic datasets takes input point clouds of size $8,192$. To put this size into perspective, it is four times of $2,048$ points accepted by PointCNN~\cite{li2018pointcnn}. {\color{\hRevise}We denote the radius of range search by $\rho$. It is set to increase by a factor of 2 between hierarchical graphs, 
while the initial $\rho$ is based on intuition. Since 0.1 is a reasonable context scale to start with, we hence apply it to all datasets except for the ShapeNet dataset. Our choice of a smaller initial $\rho$ (0.08) for ShapeNet is to benefit the learning of small parts.}
Further discussion on network configuration is also provided the related sections below. 

\subsection{ModelNet40}
The benchmark ModelNet40 dataset~\cite{wu20153d} is used to demonstrate the promise of our technique for object classification. 
The dataset comprises object meshes for 40 categories with 9,843/2,468 training/testing split. {\color{\hRevise}To train our network, we create the point clouds by uniformly sampling on mesh surfaces.} 
Compared to the existing methods  
(e.g.~\cite{ qi2017pointnet,qi2017pointnetplusplus,shen2018mining,simonovsky2017dynamic}), the convolutions performed in our network enable processing large input point clouds. 
Hence, our network is trained employing 10K input points.
The channel settings of the first MLP and the six SPH3D layers is 32 and 64-64-64-128-128-128. 
We use the same classifier 512-256-40 as the previous works~\cite{qi2017pointnet,shen2018mining,lei2019octree}. The Encoder4 in Table~\ref{tab:net_config} indicates that the network learns a global representation of the point cloud using G-SPH3D. For that, we create a virtual vertex whose associated coordinates are computed as the average coordinates of the real vertices in the graph. We connect all the real vertices to the virtual vertex, and use a spherical kernel of size $8\times2\times1+1$ for feature computation. G-SPH3D computes the feature only at the virtual vertex, that becomes the global representation of point cloud for the classifier.
\begin{table}[t]
\centering
\caption{ModelNet40 classification: Average class and instance accuracies are reported along with the number of input points per sample (\#point), the number of network parameters (\#params), and the train/test time.}\label{tab:classify}
 \begin{adjustbox}{width=0.48\textwidth}
{\LARGE\begin{tabular}{l|c|c|c|c|c|c}
  \hline
  \multirow{2}{*}{Method}
  &\multirow{2}{*}{\#point} &\multirow{2}{*}{\#params} &\multirow{2}{*}{class} & \multirow{2}{*}{instance} &
  \multicolumn{2}{c}{time(ms)} \\
  \cline{6-7}
 & & & & & training & testing \\
  \hline
  ECC \cite{simonovsky2017dynamic}&1000&0.2M& 83.2&87.4&--&--\\
  PointNet \cite{qi2017pointnet}&1024&3.5M& 86.2 &89.2 &7.9&2.5\\
  PointNet++ \cite{qi2017pointnetplusplus}&1024&1.5M&  88.0 &90.7 & 4.9 & \textbf{1.3}\\
  Kd-net(10) \cite{klokov2017escape}&1024&3.5M
  &86.3 &90.6 & --&-- \\
  SO-Net \cite{li2018so}&2048&2.4M&87.3 &90.9 &--\\
  KCNet \cite{shen2018mining}&2048&0.9M&--& 91.0& -- & --
  \\
  \color{\hRevise}DGCNN \cite{wang2018dynamic} &\color{\hRevise}1024 &\color{\hRevise}1.8M &\color{\hRevise}\textbf{89.3} &\color{\hRevise}91.5 &\color{\hRevise}15.3 &\color{\hRevise}5.6
  \\
  PointCNN \cite{li2018pointcnn}&1024& 0.6M&88.0& 91.7& 19.4&7.5\\
SFCNN \cite{rao2019spherical}&1024&8.6M&--&91.4&--&--\\
$\Psi$-CNN \cite{lei2019octree}& 10000 &3.0M& 88.7  & 92.0 &84.3&34.1 \\
  \hline
  SPH3D-GCN & 2048 & 0.7M& 88.5  & 91.4 & \textbf{4.0}&1.4 \\ 
  \cline{2-7}
 (Proposed)& 10000 & 0.8M& \textbf{89.3}  & \textbf{92.1} & 18.1&8.4 \\   
  \hline
\end{tabular}}
\end{adjustbox}
\vspace{-2mm}
\end{table}
\begin{table*}
\caption{Part Segmentation Results on ShapeNet dataset.}\label{tab:shapenet_partseg}
\vspace{-3mm}
\begin{adjustbox}{width=1\textwidth}
{\LARGE\begin{tabular}{l|c|c|c|c|c|c|c|c|c|c|c|c|c|c|c|c|c|c}
  \hline
  & instance&  class & Air- & \multirow{2}{*}{Bag} & \multirow{2}{*}{Cap} & \multirow{2}{*}{Car} & \multirow{2}{*}{Chair} & Ear- & \multirow{2}{*}{Guitar} & \multirow{2}{*}{Knife} & \multirow{2}{*}{Lamp} & \multirow{2}{*}{Laptop} & Motor- & \multirow{2}{*}{Mug} & \multirow{2}{*}{Pistol} & \multirow{2}{*}{Rocket} & Skate- & \multirow{2}{*}{Table} \\
 & mIoU&  mIoU & plane &  & & & &phone & & &  & &bike & & &  & board &  \\
 \hline
 \# shapes & & &2690& 76& 55& 898& 3758 &69 &787 &392 &1547 &451 &202 &184 &283 &66 &152 &5271\\
  \hline
  3D-CNN \cite{qi2017pointnet}& 79.4&74.9 & 75.1 &72.8 &73.3 &70.0 &87.2 &63.5 &88.4 &79.6 &74.4 &93.9 &58.7 &91.8 &76.4 &51.2 &65.3 &77.1\\
  \hline
  Kd-net \cite{klokov2017escape}
& 82.3&77.4 &80.1 &74.6 &74.3 &70.3 &88.6 &73.5 &90.2 &87.2 &81.0 &94.9 &57.4 &86.7 &78.1 &51.8 &69.9 &80.3 \\ 
  PointNet \cite{qi2017pointnet}&  83.7&80.4 &83.4 &78.7 &82.5 &74.9 &89.6 &73.0 &91.5 &85.9 &80.8 &95.3 &65.2 &93.0 &81.2 &57.9 &72.8 &80.6 \\
  Spec-CNN \cite{yi2017syncspeccnn} & 84.7&82.0 &81.6 &81.7 &81.9 &75.2 &90.2 &74.9 &\textbf{93.0} &86.1 &84.7 &95.6 &66.7 &92.7 &81.6 &60.6 &\textbf{82.9} &82.1\\
  SPLATNet$_{\text{3D}}$ \cite{su2018splatnet} &84.6 &82.0 &81.9 &83.9 &88.6 &79.5 &90.1 &73.5& 91.3& 84.7& 84.5 &96.3& 69.7& 95.0 &81.7& 59.2& 70.4& 81.3\\
  KCNet \cite{shen2018mining}& 84.7&82.2 &82.8 &81.5 &86.4 &77.6& 90.3 &76.8 &91.0 &87.2 &84.5 &95.5 &69.2 &94.4 &81.6 &60.1 &75.2 &81.3\\ 
 SO-Net \cite{li2018so} & 84.9&81.0 &82.8 &77.8 &88.0 &77.3 &90.6 &73.5 &90.7 &83.9 &82.8 &94.8 &69.1 &94.2 &80.9 &53.1 &72.9 &83.0 \\ 
 PointNet++ \cite{qi2017pointnetplusplus} & 85.1&81.9 &82.4 &79.0 &87.7 &77.3 &90.8 &71.8 &91.0 &85.9 &83.7 &95.3 &71.6 &94.1 &81.3 &58.7 &76.4 &82.6 \\
 \color{\hRevise}DGCNN \cite{wang2018dynamic}& \color{\hRevise}85.2 &\color{\hRevise}82.3 & \color{\hRevise}84.0& \color{\hRevise}83.4& \color{\hRevise}86.7& \color{\hRevise}77.8& \color{\hRevise}90.6& \color{\hRevise}74.7& \color{\hRevise}91.2& \color{\hRevise}87.5& \color{\hRevise}82.8& \color{\hRevise}95.7& \color{\hRevise}66.3& \color{\hRevise}94.9& \color{\hRevise}81.1& \color{\hRevise}63.5& \color{\hRevise}74.5& \color{\hRevise}82.6\\
  SpiderCNN
  \cite{xu2018spidercnn}&85.3&81.7 &83.5 &81.0 &87.2 &77.5 &90.7& 76.8& 91.1 &87.3 &83.3 &95.8 &70.2 &93.5 &82.7 &59.7 &75.8 &82.8\\
  SFCNN \cite{rao2019spherical}&85.4&82.7& 83.0& 83.4 &87.0& 80.2& 90.1 &75.9& 91.1& 86.2& 84.2 &96.7& 69.5& 94.8& 82.5& 59.9& 75.1& 82.9\\
  PointCNN \cite{li2018pointcnn} &86.1&84.6 &84.1 &\textbf{86.5}& 86.0& 80.8& 90.6& \textbf{79.7}& 92.3&\textbf{88.4}& 85.3& 96.1& 77.2& 95.3& 84.2& \textbf{64.2}& 80.0& 82.3\\
$\Psi$-CNN \cite{lei2019octree}& 
86.8&83.4
 &84.2 &82.1 &83.8 &80.5 &91.0 &78.3 &91.6 &86.7 &84.7 &95.6 & 74.8 &94.5 & 83.4 & 61.3 &75.9 &\textbf{85.9}\\
  \hline  
 SPH3D-GCN(Prop.) 
 &\textbf{86.8}&\textbf{84.9 }
&\textbf{84.4} &86.2 &\textbf{89.2} &\textbf{81.2}     &\textbf{91.5} &77.4 &92.5 &88.2 &\textbf{85.7} &\textbf{96.7} & \textbf{78.6} &\textbf{95.6} & \textbf{84.7} & 63.9 &78.5 &84.0\\
  \hline
\end{tabular}}
\end{adjustbox}
\vspace{-3mm}
\end{table*}

Following our preliminary work for $\Psi$-CNN \cite{lei2019octree}, we boost performance of the classification network by applying max pooling to the intermediate layers, i.e. Encoder1, Encoder2, Encoder3. We concatenate these max-pooled features to the global feature representation in Encoder4 to form a more effective representation. This results in features with $832=64+128+128+512$ channels for the classifier.
We use weight decay of $10^{-5}$ in the end-to-end network training, where  0.5 dropout~\cite{srivastava2014dropout} is also applied to the fully connected layers of the classifier to alleviate overfitting.

Table~\ref{tab:classify} benchmarks the performance of our technique that is abbreviated as SPH3D-GCN. All the tabulated techniques uses $xyz$ coordinates as the raw input features. {\color{\hRevise}We also report the training and inference time of PointNet\footnote{\href{https://github.com/charlesq34/pointnet}{https://github.com/charlesq34/pointnet.}}, PointNet++\footnote{\href{https://github.com/charlesq34/pointnet2}{https://github.com/charlesq34/pointnet2.}}, DGCNN\footnote{https://github.com/WangYueFt/dgcnn/tree/master/tensorflow.}, $\Psi$-CNN and SPH3D-GCN on our local Titan Xp GPU.} The timings for PointCNN are taken from  \cite{li2018pointcnn}, which are based on a more powerful Tesla P100 GPU. Titan Xp and Tesla P100 performance can be compared using \cite{titanxp_teslap100_theory,titanxp_teslap100_run}. {\color{\hRevise} 
These timings are computed by dividing per batch inference time for each network with the used batch size, where optimized batch sizes reported in the original works are employed.}
As shown in the Table, SPH3D-GCN and $\Psi$-CNN - our preliminary work - achieve very competitive results. 
Comparing the computational and memory advantage of SPH3D-GCN over $\Psi$-CNN, for 10K input points, SPH3D-GCN requires much less parameters (0.78M vs. 3.0M) and performs much faster (18.1/8.4ms vs. 84.3/34.1ms). 
We also report the performance of SPH3D-GCN for $2,048$ points, where the training/inference time becomes comparable to PointNet++, but performance does not deteriorates much. 
It is worth mentioning that, relative to PointCNN, the slightly higher number of parameters for our technique results from the classifier.  In fact, our parameter size for learning the global feature representation is 0.2M, which is much less than the 0.5M for the PointCNN.


\subsection{ShapeNet}
The ShapeNet part segmentation dataset \cite{yi2016scalable} contains 16,881 synthetic models from 16 categories. The models in each category have two to five annotated parts, amounting to 50 parts in total. The point clouds are created with uniform sampling from well-aligned 3D meshes.
This dataset provides $xyz$ coordinates of the points as raw features, and has 14,007/2,874 training/testing split defined. 
Following the existing works \cite{wu2014interactive,yi2016scalable,lei2019octree}, we train independent networks to segment the parts of each category. 
The configuration of our U-shape graph network is shown in Table~\ref{tab:net_config}. 
The output class number $C$ of the classifier is determined by the number of parts in each category. We standardize the input models of ShapeNet by normalizing the input point
clouds to unit sphere with zero mean. Among other ground truth labelling issues pointed out by the existing works \cite{lei2019octree,shen2018mining,su2018splatnet}, there are some samples in the dataset that contain parts represented with only one point. 
Differentiating these point with only geometric information is misleading for deep models, both from training and testing perspective. 
Hence, after normalizing each model, we also remove such points from the point cloud\footnote{{\color{\hRevise}We remove parts represented with a single point within a sphere of radius 0.3.}}. 

In Table \ref{tab:shapenet_partseg}, we compare our results with the popular techniques that also take irregular point clouds as input, using the part-averaged IoU (mIoU) metric proposed in \cite{qi2017pointnet}. 
In the table, techniques like PointNet, PointNet++, SO-Net also exploit normals besides point coordinates as the input features, which is not the case for the proposed SPH3D-GCN. In our experiments, SPH3D-GCN not only achieves the same instance mIoU as $\Psi$-CNN~\cite{lei2019octree}, but also outperforms the other approaches on 9 out of  16 categories, resulting in  the highest class mIoU 84.9\%. We also trained a single network with the configuration shown in Table~\ref{tab:net_config} to segment the 50 parts of all categories together.
In that case, the obtained instance and class mIoUs are  85.4\% and 82.7\%, respectively. These results are very close to highly competitive method SFCNN \cite{rao2019spherical}.
In all segmentation experiments, we apply the random sampling operation multiple times to ensure that every point in the test set is evaluated. 
\vspace{-2mm}
\subsection{RueMonge2014}
\begin{table}
\centering
\caption{Semantic Segmentation on RueMonge2014 dataset.}\label{ruemonge2014_seg}\label{tab:RueMonge2014_seg}
\begin{adjustbox}{width=0.45\textwidth}
\begin{tabular}{l|c|c|c}
  \hline
 Method& mAcc & OA & mIoU \\
  \hline
Riemenschneider et al. \cite{riemenschneider2014learning} & -- & -- & 42.3 \\
Martinovic et al. \cite{martinovic20153d} & -- & -- & 52.2 \\
Gadde et al. \cite{gadde2018efficient} & 68.5 &78.6 &54.4 \\
OctNet $256^3$ \cite{riegler2017octnet} & 73.6 & 81.5 & 59.2 \\
SPLATNet$_\text{3D}$ \cite{su2018splatnet} & -- & -- & 65.4\\
$\Psi$-CNN \cite{lei2019octree} & 74.7 & 83.5 & 63.6 \\
  \hline
SPH3D-GCN (Proposed) & \textbf{80.0} & \textbf{84.4} & \textbf{66.3} \\
\hline
\end{tabular}
\end{adjustbox}
\vspace{-3mm}
\end{table}
We test our technique for semantic segmentation of the real-world outdoor scenes using RueMonge2014 dataset~\cite{riemenschneider2014learning}.
This dataset contains 700 meters Haussmanian style facades along a European street annotated with point-wise labelling. There are 7 classes in total, which include \emph{window, wall, balcony, door, roof, sky} and \emph{shop}. The point clouds are provided with normals and color features. We use
$xyz$ coordinates as well as normals and color values to form 9-dim input features for a point. The detailed network configuration used in this experiment is shown in Table \ref{tab:net_config}, for which $C = 7$ for  RueMonge2014. The original point clouds are split into smaller point cloud blocks following the \textit{pcl\_split.mat} indexing file provided with the dataset. We randomly sample $8,192$ points from each block and use the sampled point clouds for training and testing. To standardize the points, we force their $x$ and $y$ dimensions to have zero mean values, and the $z$ dimension is kept non-negative. 
In the real-world applications (here and the following sections), we use data augmentation but no weight decay or dropout.
As compared to the preliminary work~\cite{lei2019octree}, we do not perform pre-processing in terms of alignment of the facade plane and gravitational axis correction. Besides, the processed blocks are also mostly much larger.
Under the evaluation protocol of \cite{gadde2018efficient}, Table~\ref{tab:RueMonge2014_seg} compares our current approach SPH3D-GCN with the recent methods, including  $\Psi$-CNN~\cite{lei2019octree}. 
It can be seen that SPH3D-GCN achieves very competitive performance, using only 0.4M parameters.
\vspace{-3mm}
\begin{table*}[!t]
    \centering
    \caption{{\color{\hReviseMinor}3D semantic labelling on Scannet: All the techniques use 3D coordinates and color values as input features for network training.}}
    \label{tab:scannet_seg}
    \begin{adjustbox}{width=1\textwidth}
    {\Huge\begin{tabular}{l|c|cccccccccccccccccccc}
    \hline
    Method & mIoU & floor &wall &chair &sofa &table& door& cab& bed &desk &toil &sink &wind& pic &bkshf &curt &show &cntr &fridg& bath &other\\
    \hline
    \hline
SSC-UNet \cite{graham2017submanifold}$^{\dagger}$ &30.8&90.8&68.5&55.3&36.3&34.5&14.8&16.6&29.0&14.7&54.6&35.4&27.8&6.4&27.8&28.6&1.8&16.9&2.3&35.3&18.2 \\
SparseConvNet \cite{graham20183d}$^{\dagger}$ &72.5&95.5&86.5&86.9&82.3&62.8&61.4&72.1&82.1&60.3&93.4&72.4&68.3&32.5&84.6&75.4&87.0&53.3&71.0&64.7&57.2 \\
 \hline
 \hline
ScanNet \cite{dai2017scannet}&30.6&78.6&43.7&52.4&34.8&30.0&18.9&31.1&36.6&34.2&46.0&31.8&18.2&10.2&50.1&0.2&15.2&21.1&24.5&20.3&14.5\\
PointNet++ \cite{qi2017pointnetplusplus}&33.9&67.7&52.3&36.0&34.6&23.2&26.1&25.6&47.8&27.8&54.8&36.4&25.2&11.7&45.8&24.7&14.5&25.0&21.2&58.4&18.3\\
  SPLATNET$_{\text{3D}}$ \cite{su2018splatnet}& 39.3&92.7&69.9&65.6&51.0&38.3&19.7&31.1&51.1&32.8&59.3&27.1&26.7&0.0&60.6&40.5&24.9&24.5&0.1&47.2&22.7\\
   Tangent-Conv \cite{tatarchenko2018tangent}& 43.8&91.8&63.3&64.5&56.2&42.7&27.9&36.9&64.6&28.2&61.9&48.7&35.2&14.7&47.4&25.8&29.4&35.3&28.3&43.7&29.8\\
  PointCNN \cite{li2018pointcnn} &45.8&94.4&70.9&71.5&54.5&45.6&31.9&32.1&61.1&32.8&75.5&48.4&47.5&16.4&35.6&37.6&22.9&29.9&21.6&57.7&28.5\\
 PointConv \cite{wu2019pointconv}& 55.6&\textbf{94.4}&76.2&73.9&63.9&50.5&44.5&47.2&64.0&41.8&82.7&54.0&51.5&\textbf{18.5}&\textbf{57.4}&43.3&57.5&\textbf{43.0}&46.4&63.6&37.2\\
 \hline
SPH3D-GCN (Prop.)& \textbf{61.0}&93.5&\textbf{77.3}&\textbf{79.2}&\textbf{70.5}&\textbf{54.9}&\textbf{50.7}&\textbf{53.2}&\textbf{77.2}&\textbf{57.0}&\textbf{85.9}&\textbf{60.2}&\textbf{53.4}&4.6&48.9&\textbf{64.3}&\textbf{70.2}&40.4&\textbf{51.0}&\textbf{85.8}&\textbf{41.4}\\
\hline
\multicolumn{22}{l}{\color{\hRevise}$^{\dagger}${We include \cite{graham2017submanifold, graham20183d} as representative 3D-CNN methods for reference only. These methods do not employ GCN.} }
    \end{tabular}}
    \end{adjustbox}
\end{table*}
\begin{table*}[!t]
\caption{Performance on S3DIS dataset: Area 5 (top), all 6 folds (bottom). 
For the Area 5, SPH3D-GCN (9-dim) follows PointNet~\cite{qi2017pointnet} to construct 9-dim input feature instead of 6-dim feature used by the proposed network.}\label{tab:s3dis_seg}
\begin{adjustbox}{width=1\textwidth}
{\LARGE\begin{tabular}{c|l|ccc|ccccccccccccc}
  \hline
 &Methods& OA& mAcc & mIoU & ceiling & floor & wall & beam & column & window & door & table & chair & sofa & bookcase & board & clutter \\
  \hline
\multirow{7}{*}{\rotatebox[origin=c]{90}{Area 5}}
& PointNet \cite{qi2017pointnet}& -- &49.0 &41.1 &88.8 &97.3 &69.8 &0.1 &3.9 &46.3 &10.8 &58.9 &52.6  &5.9 &40.3  &26.4 &33.2\\
&SEGCloud \cite{tchapmi2017segcloud} & -- &57.4 &48.9 &90.1 &96.1 &69.9 &0.0 &18.4 &38.4 &23.1 &70.4 &75.9 &40.9 &58.4 &13.0 &41.6\\
&Tangent-Conv \cite{tatarchenko2018tangent}& 82.5 &62.2 &52.8 &-- &-- &-- &-- &-- &-- &-- &-- &-- &-- &-- &-- &--\\
&SPG \cite{landrieu2017large} & 86.4 &66.5 &58.0 &89.4 &96.9 &78.1 &0.0 &42.8 &48.9 &61.6&75.4 &84.7 &52.6 &69.8  &2.1 &52.2\\
&PointCNN\cite{li2018pointcnn}& 85.9& 63.9& 57.3& 92.3& 98.2 &79.4& 0.0 &17.6& 22.8& 62.1& 74.4& 80.6& 31.7& 66.7& 62.1& 56.7\\
\cline{2-18}
&SPH3D-GCN (9-dim)
& 86.6 &65.9 &58.6 &92.2 &97.2 &79.9 &0.0&32.0 &\textbf{52.2} &41.6  &76.9
&85.3&36.5&67.2  &50.7 &50.0\\
&SPH3D-GCN (Prop.) & \textbf{87.7} &65.9 &\textbf{59.5} &\textbf{93.3} &97.1 &\textbf{81.1} &0.0
&33.2 &45.8 &43.8  &\textbf{79.7}
&\textbf{86.9}
&33.2
&\textbf{71.5}  &54.1 &53.7\\
\hline
\hline
\multirow{5}{*}{\rotatebox[origin=c]{90}{All 6 Folds}}
&PointNet \cite{qi2017pointnet} &78.5& 66.2
&47.6 &88.0 &88.7 &69.3& 42.4& 23.1 &47.5 &51.6 &42.0 &54.1 &38.2 &9.6 &29.4 &35.2\\
&Engelmann et al. \cite{engelmann2017exploring} &81.1 &66.4 &49.7 &90.3 &92.1 &67.9 &44.7 &24.2 &52.3 &51.2 &47.4 &58.1 &39.0 &6.9 &30.0 &41.9\\
&\color{\hRevise}DGCNN \cite{wang2018dynamic}& \color{\hRevise}84.1 &\color{\hRevise}-- & \color{\hRevise}56.1 &\color{\hRevise}-- &\color{\hRevise}-- &\color{\hRevise}-- &\color{\hRevise}-- &\color{\hRevise}-- &\color{\hRevise}-- &\color{\hRevise}-- &\color{\hRevise}-- &\color{\hRevise}-- &\color{\hRevise}-- &\color{\hRevise}-- &\color{\hRevise}-- &\color{\hRevise}--\\
&SPG \cite{landrieu2017large} & 85.5& 73.0& 62.1& 89.9
&95.1 &76.4 &62.8 &47.1 &55.3 &68.4 &73.5 &69.2 &63.2 &45.9 &8.7 &52.9\\
&PointCNN\cite{li2018pointcnn}& 88.1& 75.6& 65.4& 94.8& 97.3& 75.8& 63.3& 51.7& 58.4& 57.2& 71.6 &69.1& 39.1& 61.2& 52.2& 58.6\\
\cline{2-18}
&SPH3D-GCN (Prop.) & \textbf{88.6} &\textbf{77.9} &\textbf{68.9} &93.3 &96.2 &\textbf{81.9} &58.6&\textbf{55.9} &55.9 &\textbf{71.7} &72.1 &\textbf{82.4} &48.5 &\textbf{64.5} &\textbf{54.8} &\textbf{60.4}\\
  \hline  
\end{tabular}}
\end{adjustbox}
\end{table*}
\subsection{ScanNet}
ScanNet \cite{dai2017scannet} is an RGB-D video dataset of  indoor environments that contains reconstructed indoor scenes with rich annotations for 3D semantic labelling. It provides $1,513$ scenes for training and $100$ scenes for testing. Researchers are required to submit their test results to an online server for performance evaluation. The dataset provides 40 class labels, while only 20 of them are used for performance evaluation. For this dataset, we keep the network configuration identical to that used for RueMonge2014, as shown in Table~\ref{tab:net_config}, where $C=21$.
To process each scene, we first downsample the point cloud with the VoxelGrid algorithm~\cite{rusu20113d} using a $3 cm$ grid.
Then, we split each scene into $1.5m\times1.5m$ blocks, padding along each side with $0.3m$ context points. The context points themselves are neither used in the loss computation nor the final prediction. Following~\cite{li2018pointcnn}, the split is only applied to the $x$ and $y$ dimensions, whereas both spatial coordinates $(x,y,z)$ and  color values $(r,g,b)$ are used as the input features.
{\color{\hRevise}Here, $(x,y,z)$ refer to the coordinates after aligning $x$ and $y$ of each block to its center, while $z$ is aligned to the bottom point of the unsplit scene such that $z\in[0,+\infty)$. We use these \emph{aligned} $xyz$ coordinates as spatial features of the network, 
which indicates that the split blocks are processed as independent point cloud samples in both the training and test stages.}
We compare our approach with 
PointConv \cite{wu2019pointconv}, PointCNN \cite{li2018pointcnn},
Tangent-Conv \cite{tatarchenko2018tangent}, SPLATNet \cite{su2018splatnet}, PointNet++ \cite{qi2017pointnetplusplus} and ScanNet \cite{dai2017scannet} in Table~\ref{tab:scannet_seg}.
These algorithms report their performance using the $xyz$ coordinates and $rgb$ values as input features similar to our method.
A common evaluation protocol is followed by all the techniques in Table~\ref{tab:scannet_seg}.
As can be noticed, SPH3D-GCN outperforms other approaches on 16 out of 20 categories, resulting in significant overall improvement in mIoU.
The low performance of our method on \emph{picture} can be attributed to the lack of rich 3D structures. We observed that the network often confuses pictures with walls. 
\subsection{S3DIS}
The Stanford large-scale 3D Indoor Spaces (S3DIS) dataset~\cite{armeni20163d} comprises  colored 3D point clouds  collected for 6 large-scale indoor areas of three different buildings using the Matterport scanner. 
The segmentation task defined on this dataset aims at labelling  
13 semantic elements, namely;  \emph{ceiling, floor, wall, beam, column, window, door, table,
chair, sofa, bookcase, board}, and \emph{clutter}. The elements that are not among the first 12,  are considered  \emph{clutter}. 
We use the same network configuration for this dataset as used for the RueMonge2014 and ScanNet, except that $C=13$ now. 
Following the convention  \cite{qi2017pointnet,tchapmi2017segcloud,landrieu2017large,li2018pointcnn}, we perform 6-fold experiment using the six areas, and explicitly experiment with the Area 5.
It is a common practice to separately  analyze performance on Area 5 because it relates to a building not covered by the other areas~\cite{tchapmi2017segcloud}.
The used evaluation metrics include the Overall Accuracy (OA), the mean Accuracy of all 13 categories (mAcc), the Intersection Over Union (IoU) for each category, and their mean (i.e.~mIoU). 

\begin{figure*}[ht]
    \centering
    \begin{tabular}{cc|cc}
    Office & mIoU$=94.4\%$ &Office &  mIoU$=91.2\%$ \\
 \includegraphics[height=36mm]{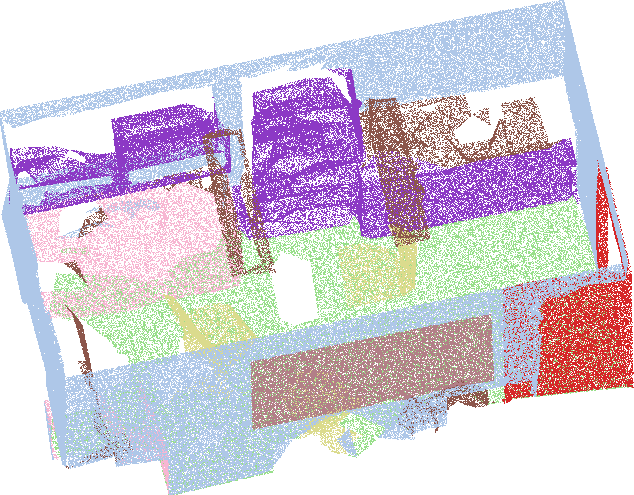}& 
 \hspace{-4mm}
 \includegraphics[height=36mm]{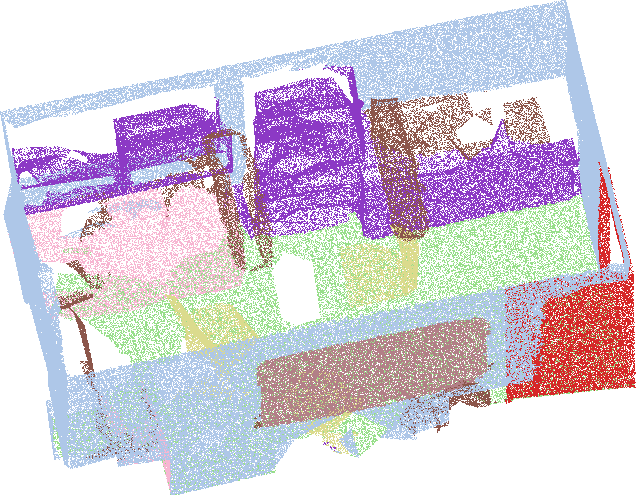}& 
 \includegraphics[height=36mm]{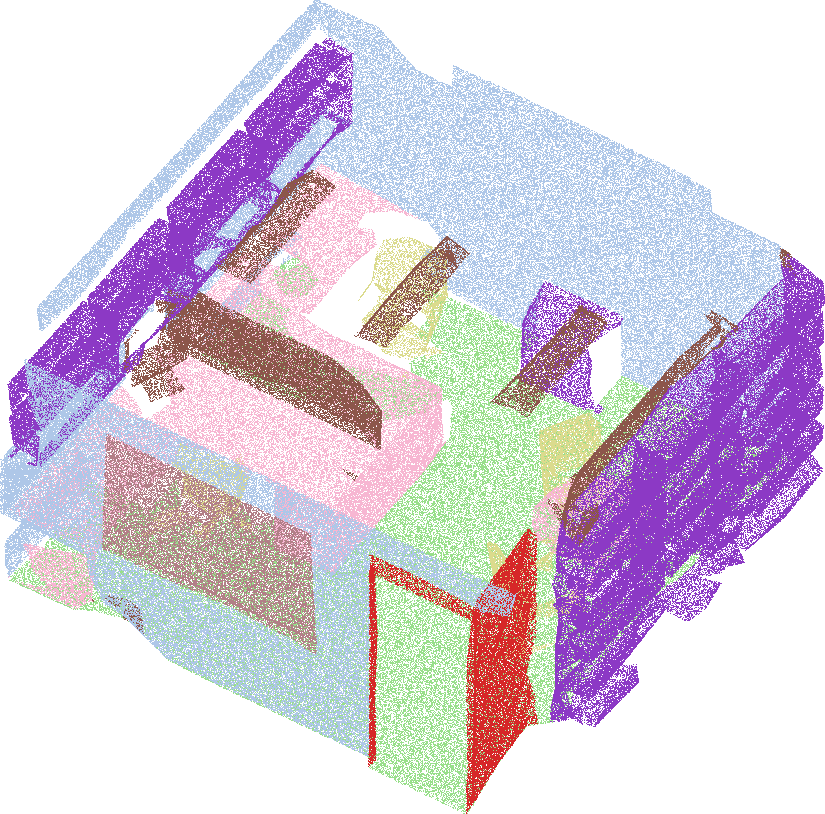}& 
  \hspace{-3mm}
 \includegraphics[height=36mm]{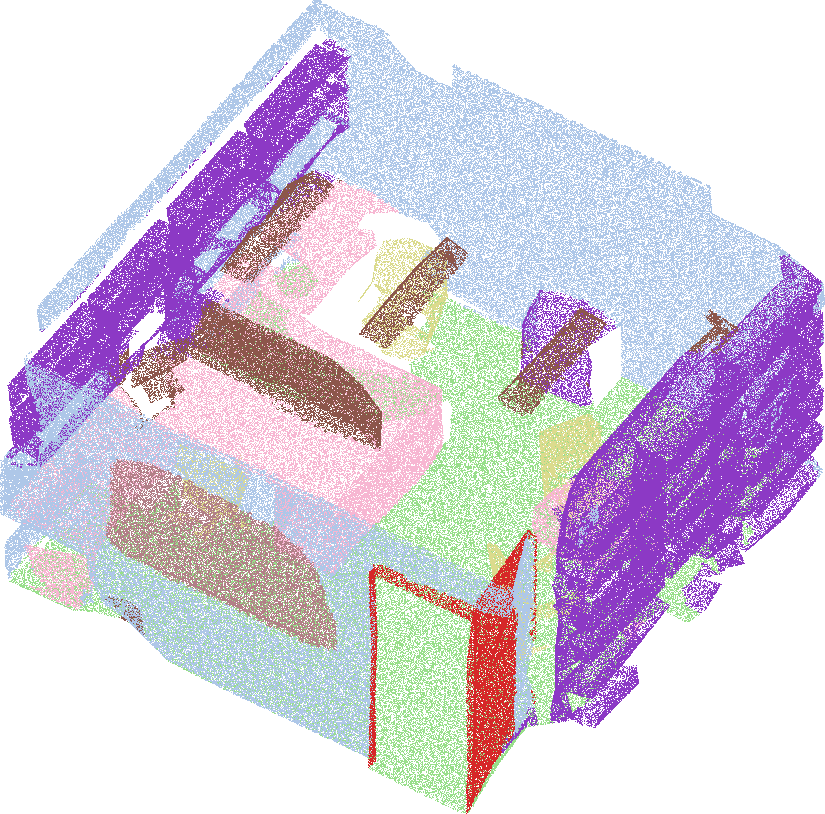}\\
  Ground truth&Proposed&Ground truth &Proposed\\
    \end{tabular}
    \caption{
    {\color{\hRevise}Prediction visualization for two representative scenes of Area 5 in S3DIS dataset. We have removed the \emph{ceiling} so that the details inside the offices are clearly visible. Despite the scene complexity, the proposed SPH3D-GCN generally segments the points accurately.}}
    \label{fig:pred_vis_s3dis}
    \vspace{-3mm}
\end{figure*}

Most of the scenes in S3DIS has millions of points. We use the same downsampling and block splitting strategy as in ScanNet. 
The input features also comprise 3D coordinates and color values that are standardized similar to those in ScanNet.  
The results of our experiments are summarized in Table~\ref{tab:s3dis_seg}. {\color{\hRevise}
Since our work focuses on deep convolutional networks for point cloud processing, 
we make comparison to the original SPG method \cite{landrieu2017large} in the table.  The SSP+SPG enhancement \cite{landrieu2019point} of \cite{landrieu2017large} mainly contributes towards over-segmentation of point clouds, which is not the topic of this paper.}
With 0.4M parameters, the proposed SPH3D-GCN achieves much better performance than the other convolutional networks (e.g. \cite{landrieu2017large,li2018pointcnn}).
For the experiments on Area 5, we also report results of an additional experiment with SPH3D-GCN(9-dim) that follows  PointNet \cite{qi2017pointnet} in creating the input feature. The 9-dim input feature comprises $xyz$+$rgb$ values and the relative location of the point in the scene.
Comparing the performance of the proposed network that uses 6-dim input feature, we notice that removing the relative locations actually benefits the performance, which can be attributed to  sensitivity of the relative locations to the scene scale. Finally, we visualize two representative prediction examples generated by our technique for the segmentation of Area 5 in Fig.~\ref{fig:pred_vis_s3dis}. As can be noticed, despite the complexity of the scenes, SPH3D-GCN is able to segment the points effectively. 

\begin{table*}[t]
    \centering
        \caption{
        {\color{\hRevise} Runtime and memory consumption comparison between regular and separable kernel. `Random' indicates random point clouds and random input features of the mentioned sizes with $C_{in} = 64$, $C_{out} =  128$, $\lambda = 2$ and batch size 16. Due to memory requirements, regular-V1 is unable to process point clouds of size 16384, whereas regular-V2 is unable to perform classification and segmentation on the standard datasets in reasonable time. Separable convolution provides considerable memory and runtime advantage to allow better scalability. We use 2048 input points for both
        classification and segmentation tasks.}}
    \label{tab:regular_separable}
   \begin{adjustbox}{width=1\textwidth}
  {\LARGE\begin{tabular}{l|c|c|c|c||c|c|c|c|c|c|c|c}
    \hline
        \multirow{4}{*}{Convolution} &  \multicolumn{4}{c||}{Random}&  \multicolumn{4}{c|}{\multirow{2}{*}{ModelNet40 (Classification)} }&\multicolumn{4}{c}{\multirow{2}{*}{ShapeNet \emph{Table} (Segmentation)}}\\
        \cline{2-5}
         &  2048 & 4096 & 8192 & 16384 & \multicolumn{4}{c|}{} & \multicolumn{4}{c}{} \\         \cline{2-13}
         \noalign{\smallskip}
         \cline{2-13}
         & \multicolumn{4}{c||}{single forward pass}  &batch& instance & GPU & train/test& batch&instance & GPU & train/test\\
         & \multicolumn{4}{c||}{per-batch time(ms)}&size&accuracy & memory & time(ms)& size& mIoU& memory & time(ms)\\
         \hline
         separable & \textbf{14.0}& \textbf{38.7}& \textbf{98.5}& \textbf{280.3}&\textbf{32}& \textbf{91.4}
        &  \textbf{0.8GB}&\textbf{4.0/1.4}&\textbf{32}&84.0&\textbf{4.4GB}&\textbf{33.4/9.4}\\
        \hline
        regular-V1 (fast) & 36.1&94.3 & 214.8& N/A&\textbf{32}& 91.3
        &8.7GB &12.5/5.3&12&\textbf{84.1}&10.7GB&180.8/25.2\\
        \hline
        regular-V2 (slow) & 84.5&453.0 &1185.1 &2746.3 &\multicolumn{8}{c}{--}\\
        \hline
    \end{tabular}}
\end{adjustbox}
\end{table*}
\begin{table*}[!h]
    \centering
    \caption{{\color{\hRevise}Ablation study (using Area 5 of S3DIS) when pooling is changed to average pooling, data augmentation is excluded and when uniform or weighted interpolation is performed. The `baseline' column has results for our proposed method that uses max pooling+with augmentation+uniform interpolation. Average pooling is not as good as max pooling and data augmentation does improve the results. Weighted interpolation works better when the number of points is high (i.e. dense point cloud) but degrades the performance in case the points are less (i.e. sparse point cloud).} 
    }
    \label{tab:s3dis_ablations}
    \begin{adjustbox}{width=0.8\textwidth}{
    \begin{tabular}{l|c|c|c|c|c|c}
    \hline
     \multirow{3}{*}{Metrics}& \multicolumn{4}{c|}{8192 points}
    & \multicolumn{2}{c}{4096 points} \\
    \cline{2-7}
    &\multirow{2}{*}{baseline}& average & without & weighted  & uniform & weighted   \\
   &  & pooling & augmentation & interpolation & interpolation & interpolation  \\
    \hline
        OA & 87.7&87.3&  86.4& 88.1&88.2&87.8\\
        \hline
        mAcc & 65.9&65.5&  64.4& 67.2&68.8&67.8\\
        \hline
        mIoU & 59.5&58.8& 57.0& 61.2&62.2&61.1\\
        \hline
    \end{tabular}}
    \end{adjustbox}
\end{table*}
\section{Ablation study}\label{sec:Ablation}
{\color{\hRevise}\noindent\textbf{Regular vs. separable:}
In contrast to regular convolution, the use of separable convolution is a major difference between the proposed technique and its preliminary work \cite{lei2019octree}. Hence, here we analyze the two choices to clarify the motivation of preferring the latter in this paper.
In the following, `$m$' denotes the input size of the point cloud, $C_{in}$ and $C_{out}$ are respectively the number of input and output feature maps of a kernel, $K$ is the maximum edge connections allowed by each vertex in the graph, and $\lambda$ is the  multiplier of the separable convolution. Let the number of kernel bins be $F$, which equals to $n\times p\times q+1$ for the spherical kernel. To this end,
the number of learnable parameters in the regular convolutional kernel is $F \times C_{in}\times C_{out}$, while the number of learnable parameters in the separable convolutional kernel is $(F\times C_{in}\times \lambda) + (C_{in}\times \lambda\times C_{out})$. 
The ratio of the former to the latter is $\frac{F\times C_{out}}{\lambda\times (F+C_{out})}$. Generally, $F\ll C_{out}$, which allows us to write $\frac{F\times C_{out}}{\lambda\times (F+C_{out})}\approx\frac{F}{\lambda}\gg1$.
Hence, the number of parameters of separable convolutional kernel is significantly smaller than that of the regular kernel.
Regarding the computational complexity, the regular convolution contains `at most' $m\times K\times C_{in}\times C_{out}$ multiplications, while the separable convolution contains at most $m\times \lambda\times C_{in}\times (K+C_{out})$ multiplications. The former is $\frac{K\times C_{out}}{\lambda\times(K+ C_{out})}$ times of the latter.  For the common setup of $K\ll C_{out}$, we have $\frac{K\times C_{out}}{\lambda\times(K+ C_{out})}\approx\frac{K}{\lambda}\gg1$.  
This indicates that the number of computations of separable convolution is also significantly less.
Reductions in both memory and computation make separable convolution a natural choice for processing large-scale point clouds using densely connected graph representations.
 
From the implementation view-point, there are two ways to achieve the regular convolution. One is to use the `tf.matmul' operation of Tensorflow which exploits the optimized CUBLAS library, referred as regular-V1 in the text to follow. The other is to wrap all the convolutional computations as bottom-level CUDA kernels, referred as regular-V2. The regular-V1 is efficient because of CUBLAS, but memory expensive because context aggregation demands padding of the input feature maps along an additional neighboring dimension.  
KPConv \cite{thomas2019kpconv} exploits `tf.matmul' for computational efficiency. However, this inevitably restrains its scalability. The regular-V2 is memory friendly, but inefficient because it fails to exploit CUBLAS. We note that separable convolution does not suffer from the implementation dilemma of the regular convolution. On one hand, it performs context aggregation with the spatial convolution on CUDA kernels in the bottom level, which saves a large amount of memory. One the other hand, for the point-wise convolution, it can readily exploit the `tf.matmul' operator without padding requirement. 
The low theoretical complexity along with efficient implementation solution make our technique scalable with separable spherical convolutions.

For quantitative comparison, we first compare the forward pass time of input samples of different sizes for separable convolution, regular-V1 and regular-V2. Fixing the batch size to 16, we generate the point clouds and their input features randomly, as we are not concerned with the accuracy at this stage. These results are summarized in the left column of Table \ref{tab:regular_separable}. Due to the memory requirements, regular-V1 is unable to process samples of size 16384 under batch size 16. We also compute the performance of our technique for classification and segmentation with both types of convolutions. These results are also included in Table \ref{tab:regular_separable}. We use ModelNet40 for classification and the \textit{Table} category of ShapeNet for segmentation. 
Due to its unreasonably slow computation, we exclude the results of regular-V2 which is expected to achieve similar accuracy as regular-V1 because both variants are actually implementations of the exact same mathematical operation.
}

\begin{table*}[t]
    \centering
    \caption{\color{\hRevise}Computational and memory requirements of the proposed technique and comparison to PointCNN. The performance is also included for reference.}
   \label{tab:s3dis_pointcnn_ours_time}
    \begin{adjustbox}{width=1\textwidth}
    {\LARGE\begin{tabular}{l|c|c|c|c|c|c|c|c|c|c|c|c|c|c}
    \hline
    \multirow{3}{*}{Method} & \multirow{3}{*}{GPU}& \multirow{3}{*}{\shortstack[c]{batch\\ \\ \\ \\ size}}&  \multirow{3}{*}{\#params}&\multirow{3}{*}{\#point}
    &\multirow{3}{*}{\shortstack[c]{GPU$^{\dagger}$\\ \\ \\ \\ memory}}& \multicolumn{6}{c|}{time(ms)}&\multicolumn{3}{c}{performance}\\
     \cline{7-15}
      & & & & & & \multicolumn{2}{c|}{graph construction} &\multicolumn{2}{c|}{training} & \multicolumn{2}{c|}{inference}&\multirow{2}{*}{OA}&\multirow{2}{*}{mAcc}&\multirow{2}{*}{mIoU}\\
     \cline{7-12}
   & & & & & &per-batch &per-sample&per-batch &per-sample &per-batch &per-sample&&& \\
     \hline
     PointCNN &Tesla P100&12&4.4M&2048& --&--&--&610&50.8& 250&20.8&85.9&63.9&57.3  \\
     \hline
     \multirow{6}{*}{SPH3D-GCN} &\multirow{6}{*}{Titan Xp}&\multirow{6}{*}{16}&\multirow{6}{*}{0.4M}&2048&2.31GB& 26 &1.6 &450&28.1&150&9.4 &87.9&67.8&61.0 \\
     \cline{5-15}
     &&&&4096&2.31GB& 35&2.2 &566&35.4&201&12.6&88.2&68.8& 62.2\\
     \cline{5-15}
      &&&&8192&4.36GB& 65 &4.1 &869&54.3&337&21.1 &87.7&65.9& 59.5\\
      \cline{5-15}
       &&&&16384&4.36GB& 162&10.1 &1509&94.3&650&40.6&87.3&66.2& 59.2\\
       \cline{5-15}
       &&&&32768&8.45GB& 511&31.9 &2803&175.2&1354&84.6&85.8&63.9&56.1
       \\
       \cline{5-15}
       &&&&65536&11.10GB& 1816&113.5 &5880&183.8&3311&103.5&--&--&--
       \\
     \hline
     \multicolumn{15}{l}{\color{\hRevise}$^{\dagger}$The GPU memory usage is computed with NVIDIA's standard `nvidia-smi' command.}
    \end{tabular}}
    \end{adjustbox}
    \vspace{-3mm}
\end{table*}
{\color{\hRevise}\noindent\textbf{Pooling \& unpooling \& augmentation:} To analyze the contribution of different constituents of our technique, we also perform ablation study by varying our  pooling  strategy, interpolation  method  and data  augmentation. In Table \ref{tab:s3dis_ablations}, we summarize the results of these experiments using the Area 5 of S3DIS. By default, our technique uses max-pooling, uniform interpolation and data augmentation, which constitutes the `baseline' in the table. Variations from the baseline are noted as the labels of the columns. We provide results for the interpolations for two different point cloud sizes, i.e.~8192 and 4096. From the table, we can conclude that 1) the adopted max pooling strategy is more beneficial than average pooling for our network. 2) Data augmentation provides a reasonable performance boost to our technique.  3) Weighted interpolation can provide better results for denser point clouds.  However, its performance against uniform interpolation is inferior for sparser clouds.}
\section{Discussion}\label{sec:Discuss}
\noindent\textbf{Scalability:} 
The combination of discrete kernel,  separable convolution and graph-based architecture adds to the scalibility of the proposed SPH3D-GCN. 
In Table~\ref{tab:s3dis_pointcnn_ours_time}, we compare our network on computational and memory grounds with a highly competitive convolutional network PointCNN that is able to take $2,048$ points as input. The reported values are for S3DIS, using the configuration in Table~\ref{tab:net_config} for our network, where we vary the input point size. {\color{\hRevise}The performance is also provided for reference. We note that smaller input sizes tend to produce slightly better results because they benefit more from data augmentation of random sampling. We use split blocks of at most 30,000 points. Hence, inputs with more than 30000 points (e.g.~32768, 65536) are affected adversely as their size increases. We provide the results for 32,768 points as a reference for such cases.}
We show the memory consumption and training/testing time of our network{\color{\hRevise}, as well as the graph construction time. From the table,  the convolution time can be computed as the difference between the inference time and graph construction time.} With a batch size 16 on 12GB GPU, our network can take point cloud of size up to $65,536$, which is identical to the number of pixels in a $256\times256$ image.
It is worth mentioning that the memory consumption of our `segmentation' network for $32,678$ input points is slightly lower than that of PointNet++ `classification' network  for $1,024$ points (8.45GB vs. 8.57GB), using the same batch size, i.e.~16. Our 0.4M parameters are  10+ times less than the 4.4M of PointCNN. Considering that we use a larger batch size than the PointCNN, we include both the per-batch and per-sample training/testing time for a fair  comparison.
It can be seen that our per-sample running time for $2,048$, $4,096$, and $8,192$ points is less than or comparable to that of PointCNN for $2,048$ points. We refer to the websites \cite{titanxp_teslap100_theory,titanxp_teslap100_run} for a speed comparison between Tesla P100 and Tian Xp. Although our  SPH3D-GCN can take larger input size, we use point cloud of size $8,192$ for S3DIS in Table \ref{tab:s3dis_seg} in the interest of time.

\begin{figure}[!t]
    \centering
    \includegraphics[width=0.48\textwidth]{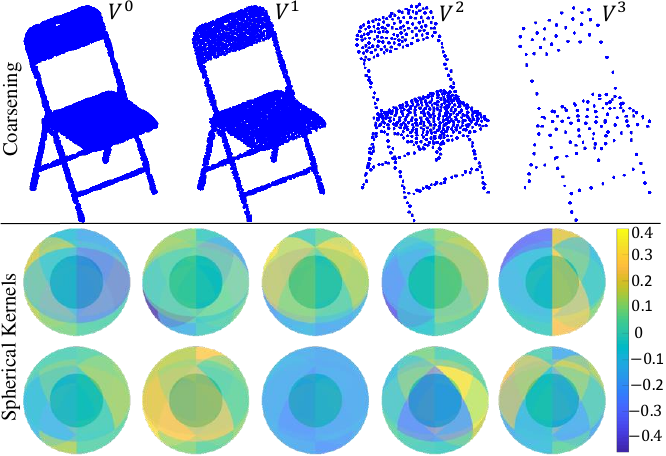}
    \caption{(Top) Graph coarsening with FPS: A chair is coarsened from left to right into point clouds of smaller resolutions. 
    (Bottom) Kernel visualization: Each row shows five spherical kernels learned in an SPH3D layer of the network for S3DIS.}
    \label{fig:FPSample_visualizedKernel}
    \vspace{-3mm}
\end{figure}

\noindent\textbf{Graph coarsening visualization:}
We coarsen point cloud along our network with the Farthest Point Sampling (FPS) that reduces graph resolution layer-by-layer, similar to the image resolution reduction in the standard CNNs. 
We visualize the coarsening effects of the FPS in Fig.~\ref{fig:FPSample_visualizedKernel}(top), using a chair from ModelNet40 as an example. The point clouds from left to right associate to the vertices of graphs $G^0, G^1, G^2, G^3$ in the network of ModelNet40.
The resolution of the point cloud systematically reduces from left to right.  
Specifically, according to Table~\ref{tab:net_config}, these point clouds contain $10$K, $2500$, $625$, $156$ points.

\noindent\textbf{Kernel visualization:}
We also visualize few learned spherical kernels in Fig.~\ref{fig:FPSample_visualizedKernel}(bottom). The two rows correspond to the spherical kernels of two SPH3D layers in Encoder2 of the network for S3DIS dataset. The size of these kernels are $8\times2\times2+1$.
As can be noticed, the weights of different kernels distribute differently in the range $[-0.5,0.4]$. For example, the third kernel in the first row 
contains positive weights dominantly in its upper hemisphere, but negative weights in the lower hemisphere, while the kernel exactly below it is mainly composed of negative weights. 
These differences indicate that different kernels can identify different features for the same neighborhoods. For better  visualization, we color each bin only on the sphere surface, not the 3D volume. Moreover, we also do not show the weight of self-loop. 

\section{Conclusion}
We introduced separable spherical convolutional kernel for point clouds and demonstrated its utility with graph pyramid architectures. We built the graph pyramids with  range search and farthest point sampling techniques. 
By applying the spherical convolution block to each graph resolution, the resulting graph convolutional networks are able to learn more effective features in larger contexts, similar to the standard CNNs. To perform the convolutions, the spherical kernel partitions its occupied space into multiple bins and associates a learnable parameter with each bin. The  parameters are learned with network training. 
We down/upsample the vertex features of different graphs with pooling/unpooling operations.
The proposed convolutional network is shown to be efficient in processing high resolution point clouds, achieving highly competitive performance on the tasks of classification and semantic segmentation on synthetic and large-scale real-world datasets.


%



\ifCLASSOPTIONcompsoc
  \section*{Acknowledgments}
  This research is supported by the Australian Research Council (ARC) grant DP190102443. The Titan Xp  GPU used for this research is donated by the NVIDIA Corporation.
\else
  \section*{Acknowledgment}
\fi


\ifCLASSOPTIONcaptionsoff
  \newpage
\fi


\bibliography{LHBib}
\bibliographystyle{IEEEtran}
\vspace{-10mm}
%
%
%

%
\begin{IEEEbiography}[{\includegraphics[width=1in,height=1.25in,clip,keepaspectratio]{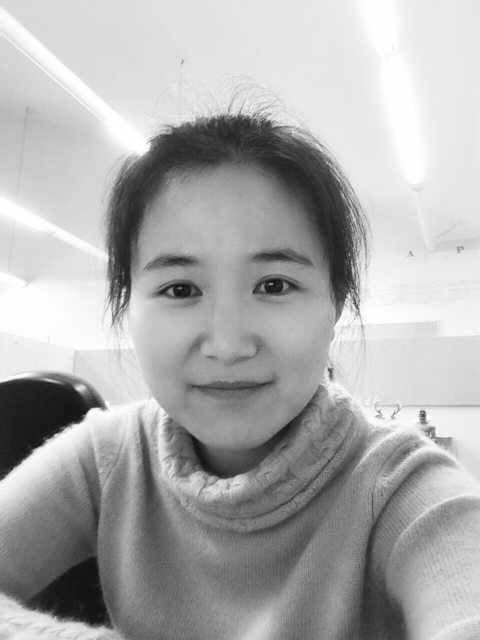}}]
{Huan Lei}
is currently working toward the
Ph.D. degree at The University of Western Australia (UWA) under the supervision of  Prof. Ajmal Mian and Dr. Naveed Akhtar. She is a recipient
of the UWA International Postgraduate Research Scholarship.
Her research interests are geometric deep learning and their applications in 3D vision. Her research interests include graphical models, relational networks, adversarial machine learning and medical image processing.
\end{IEEEbiography}
\vspace{-10mm}
\begin{IEEEbiography}[{\includegraphics[width=1in,height=1.25in,clip,keepaspectratio]{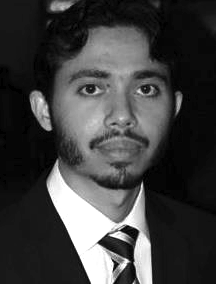}}]
{Naveed Akhtar}
received his PhD in Computer Vision from The University of Western Australia (UWA) and Master degree in Computer Science from Hochschule Bonn-Rhein-Sieg, Germany (HBRS). His research in  Computer Vision and Pattern Recognition has regularly published in  prestigious venues of the field, including IEEE CVPR and IEEE TPAMI. He is also serving as an Associate Editor of IEEE Access. During his PhD, he was recipient of multiple scholarships, and winner of the Canon Extreme Imaging Competition in 2015. Currently, he is a Lecturer at UWA since July 2019. Previously, he has also served as a Research Fellow and a Lead Research Fellow in Computer Vision at UWA and the Australian National University for three years. His current research interests include action recognition, adversarial machine learning, 3D point clouds and hyper-spectral image analysis.
\end{IEEEbiography}
\vspace{-10mm}
\begin{IEEEbiography}[{\includegraphics[width=1in,height=1.25in,clip,keepaspectratio]{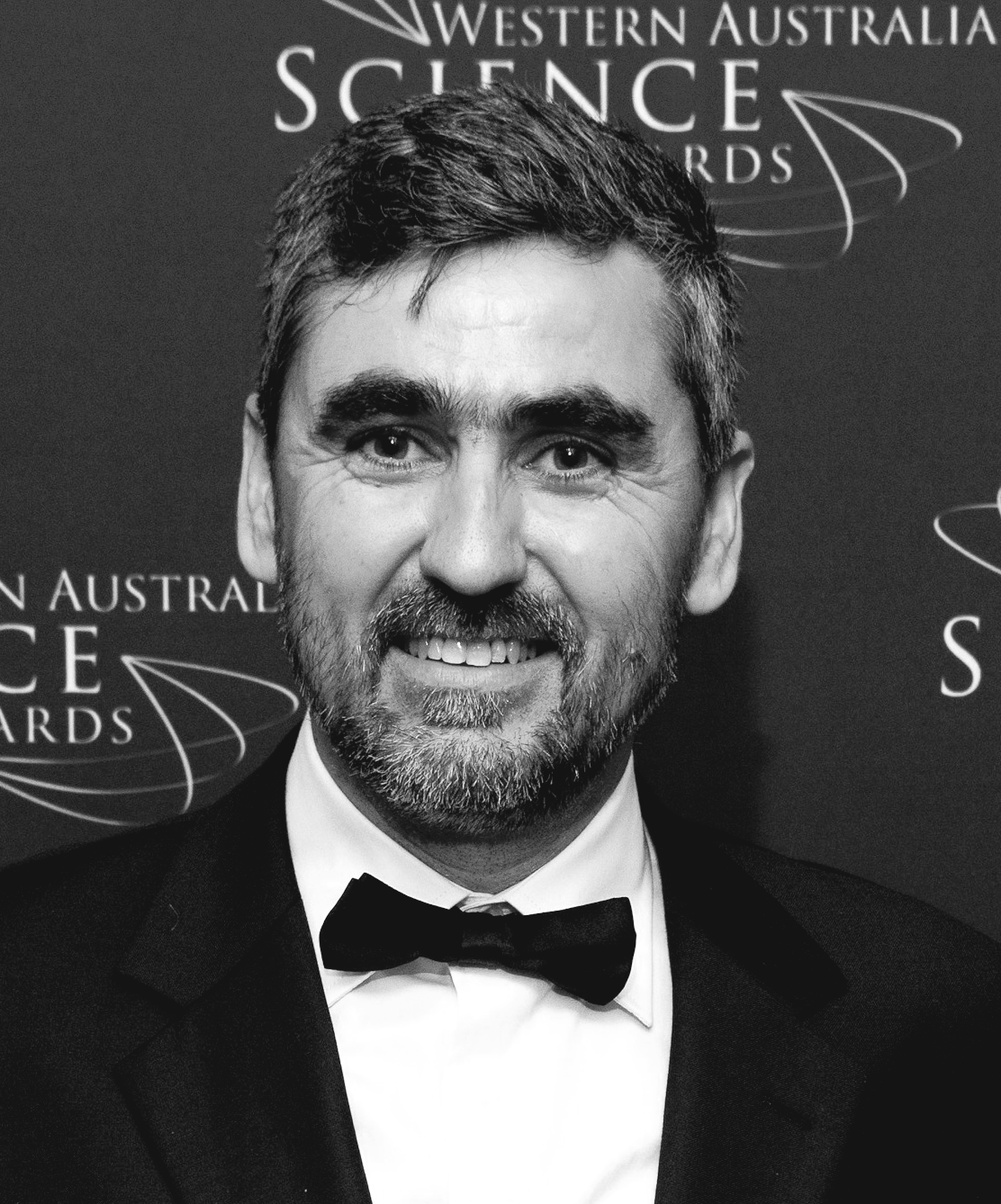}}]
{Ajmal Mian}
is a Professor of Computer Science at The University of Western Australia. He has received two prestigious fellowships and several research grants from the Australian Research Council and the National Health and Medical Research Council of Australia with a combined funding of over \$12 million. He was the West Australian Early Career Scientist of the Year 2012 and has received several awards including the Excellence in Research Supervision Award, EH Thompson Award, ASPIRE Professional Development Award, Vice-chancellors Mid-career Award, Outstanding Young Investigator Award, the Australasian Distinguished Dissertation Award and various best paper awards. His research interests are in computer vision, 3D deep learning, shape analysis, face recognition, human action recognition and video analysis.
\end{IEEEbiography}




\end{document}